# MHub.ai: A Simple, Standardized, and Reproducible Platform for AI Models in Medical Imaging

Leonard Nürnberg[1,2,3], Dennis Bontempi[1,2,3], Suraj Pai[1,2,3], Curtis Lisle[4], Steve Pieper[5], Ron Kikinis[6], Sil van de Leemput[7], Rahul Soni[8], Gowtham Murugesan[8], Cosmin Ciausu[1,2,3], Miriam Groeneveld[7], Felix J. Dorfner[9,10], Jue Jiang[11], Aneesh Rangnekar[11,], Harini Veeraraghavan[11], Joeran S. Bosma[7], Keno Bressem[12,13], Raymond Mak[1,3], Andrey Fedorov[6], Hugo JWL Aerts[1,2,3]

[1] Artificial Intelligence in Medicine (AIM) Program, Mass General Brigham, Harvard Medical School, Boston, MA, USA.
[2] Radiology and Nuclear Medicine, CARIM & GROW, Maastricht University, Maastricht, Netherlands.
[3] Department of Radiation Oncology, Dana-Farber Cancer Institute, Brigham and Women's Hospital, Harvard Medical School, Boston, MA, USA.
[4] KnowledgeVis, LLC, Maitland, FL, USA.
[5] Isomics, Inc., Cambridge, MA, USA
[6] Department of Radiology, Brigham and Women's Hospital, Harvard Medical School, Boston, MA, USA.
[7] Imaging Department, Radboud University Medical Center, Nijmegen, The Netherlands
[8] BAMF Health LLC, Grand Rapids, MI, USA
[9] Department of Radiology, Charité - Universitätsmedizin Berlin corporate member of Freie Universität Berlin and Humboldt Universität zu Berlin, Berlin, Germany
[10] Athinoula A. Martinos Center for Biomedical Imaging, Massachusetts General Hospital and Harvard Medical School, 149 Thirteenth St, Charlestown, MA, USA
[11] Department of Medical Physics, Memorial Sloan Kettering Cancer Center, New York, NY, USA
[12] Department of Diagnostic and Interventional Radiology, Technical University of Munich, School of Medicine and Health, Klinikum rechts der Isar, TUM University Hospital
[13] Department of Cardiovascular Radiology and Nuclear Medicine, Technical University of Munich, School of Medicine and Health, German Heart Center, TUM University Hospital

**ABSTRACT**

Artificial intelligence (AI) has the potential to transform medical imaging by automating image analysis and accelerating clinical research. However, research and clinical use are limited by the wide variety of AI implementations and architectures, inconsistent documentation, and reproducibility issues. Here, we introduce MHub.ai, an open-source, container-based platform that standardizes access to AI models with minimal configuration, promoting accessibility and reproducibility in medical imaging. MHub.ai packages models from peer-reviewed publications into standardized containers that support direct processing of DICOM and other formats, provide a unified application interface, and embed structured metadata. Each model is accompanied by publicly available reference data that can be used to confirm model operation. MHub.ai includes an initial set of state-of-the-art segmentation, prediction, and feature extraction models for different modalities. The modular framework enables adaptation of any model and supports community contributions. We demonstrate the utility of the platform in a clinical use case through comparative evaluation of lung segmentation models. To further strengthen transparency and reproducibility, we publicly release the generated segmentations and evaluation metrics and provide interactive dashboards that allow readers to inspect individual cases and reproduce or extend our analysis. By simplifying model use, MHub.ai enables side-by-side benchmarking with identical execution commands and standardized outputs, and lowers the barrier to clinical translation.

# INTRODUCTION

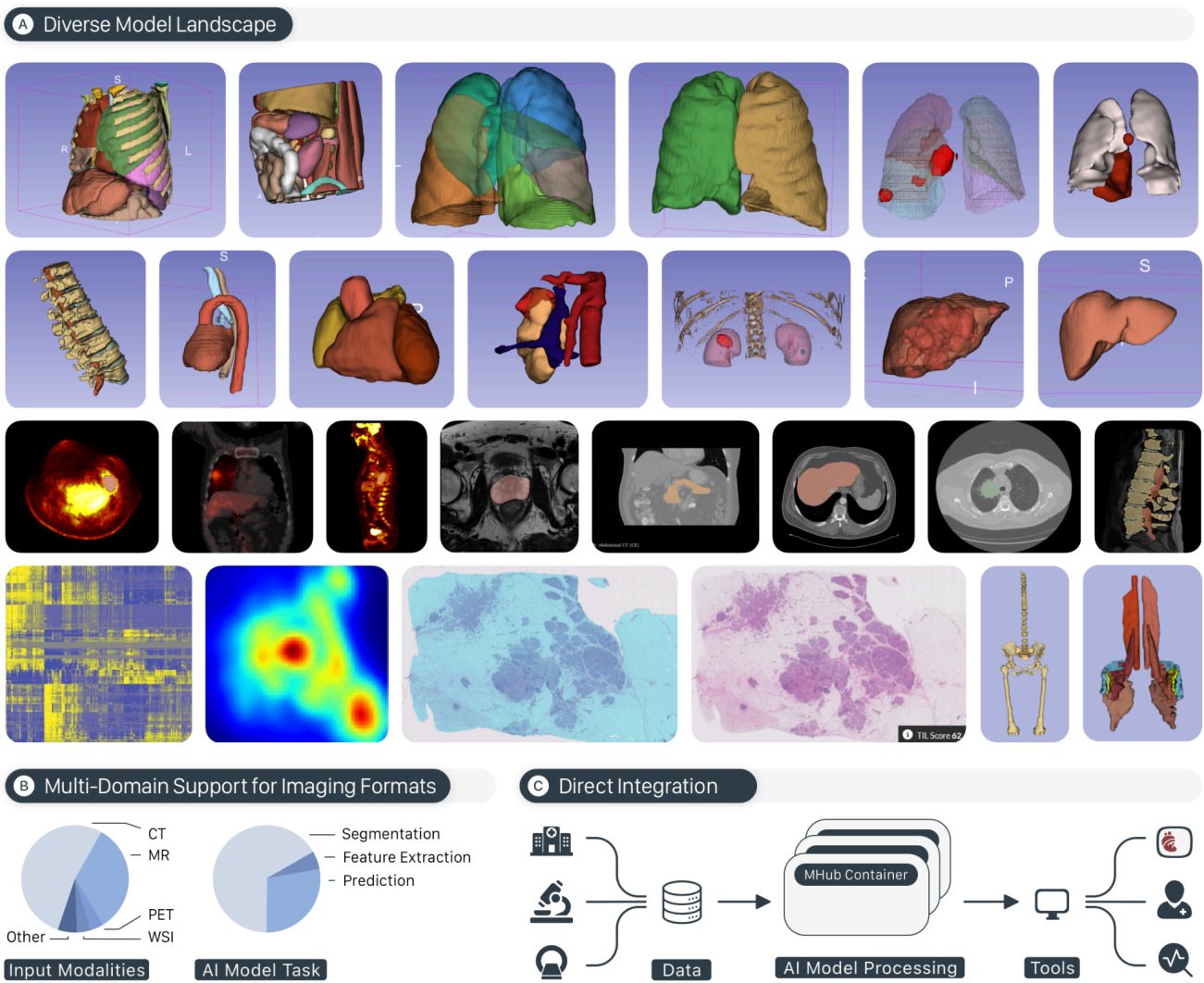

**Figure 1** a) *Diverse model landscape:* Representative examples of the models included in the MHub.ai platform for various imaging modalities, domains, and tasks. More than half of the models overlap with others in one or more outputs, increasing the available choices. b) *Multi-domain support for imaging formats:* The framework supports several input modalities, such as Computed Tomography (CT), Magnetic Resonance Imaging (MRI), Positron Emission Tomography (PET), Whole Slide Imaging (WSI), X-ray, and image formats such as Joint Photographic Experts Group (JPG) or Portable Network Graphics (PNG), and can be extended to support additional modalities and file formats. Segmentation, prediction, and feature extraction models are supported by the framework and are represented in the model repository. c) *Direct integration:* The MHub.ai framework adapts existing AI pipelines to the clinical imaging standard format DICOM, enabling direct integration with existing public and private data providers as well as third-party tools for visualization, inspection, organization, and analysis.

Artificial Intelligence (AI) methods, such as convolutional neural networks, have shown remarkable capabilities in interpreting medical imaging in radiology, pathology, and beyond (Hosny et al., 2018; Niazi et al., 2019; Huynh et al., 2020; Kann et al., 2021;McGenity et al., 2024; Mienye et al., 2025). Multiple studies have shown that these models can perform tasks such as segmentation, classification, and feature extraction with an accuracy comparable to human experts  (Liu et al., 2019; Aggarwal et al., 2021). However, many challenges remain in translating these advances into real-world clinical and scientific workflows (Korfiatis et al., 2025; Xu et al., 2025; Bontempi et al., 2024; Seneviratne et al., 2019).

Key barriers include inconsistent model implementations, complex installation procedures, stringent data pre-processing requirements, unclear configuration parameters, and arbitrary output formats - challenges often compounded by incomplete or inadequate documentation (Koçak et al., 2024; Theriault-Lauzier et al., 2024; Beam et al., 2020). Furthermore, while the file format standard used in hospitals and clinics is DICOM (Clunie et al., 2000; Mildenberger et al., 2002), AI models are typically trained and operate on alternative formats, which require manual and often time-consuming conversion and organization. Together, these factors create substantial friction for other investigators, hinder model comparability, and undermine reproducibility. Consequently, most published AI models remain unused by external researchers and are rarely translated into clinical practice (Seneviratne et al., 2019; Najjar, 2023).

Here, we introduce MHub.ai, a platform that provides containerized AI models in medical imaging with a standardized application interface, built in DICOM compatibility and reproducibility tests. The model repository offers a wide variety of public medical imaging AI models for different input modalities, imaging domains, and tasks (see **Figure 1a, b**). The underlying framework provides tools to unify the application interface across models and enables direct integration with existing data providers, data organization, and visualization tools by adapting existing standards (**Figure 1c**). By aligning AI model inference and providing automated reproducibility tests, MHub.ai reduces technical friction for model execution and enables third-party investigators to consistently use, evaluate, and compare models, supporting systematic benchmarking and lowering barriers to clinical adoption.

First, we describe the platform architecture and framework, followed by an overview of the model repository, which provides public models that can be used directly. The utility of MHub.ai is illustrated by comparing three lung segmentation models publicly available on the platform using public clinical data.

## RESULTS

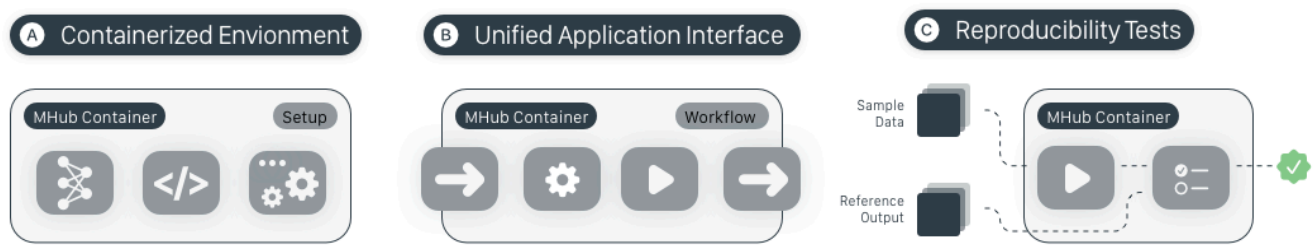

**Figure 2** *a) Containerized models*. Each AI model is packaged in an isolated container that includes inference code, pre-trained weights, dependencies, libraries, and the runtime environment for platform-independent execution with no individual setup required. *b) Unified application interface*. A standardized interface provides native DICOM handling, workflow orchestration, and consistent input and output structures across models. *c) Reproducibility tests*. Public sample data and reference outputs are provided for every model. The framework's test engine runs each model on specific sample data, compares the outputs with the reference data, and reports the test results in a human- and machine-readable format.

**System Design and Architectural Overview:** MHub.ai is implemented as a modular platform comprising three primary components (see **Figure 2**). The first component is the *Containerized Environment*, where each model is packaged on top of the MHub.ai base image and bundles the inference source code, the model weights, the dependencies, and the runtime environment (**Figure 2a**). Containerization enables portability, cross-platform compatibility and isolation, while base layer sharing reduces disk footprint.

The second component is the *Unified Application Interface* used by all containers, which harmonizes the inputs and outputs (IO), configuration, and execution of the models. All containers directly accept clinical images in the native DICOM format and others, and provide outputs in a per-output type standardized representation (e.g. DICOM-SEG for segmentations, JSON/CSV for tabular data), referencing the input images unique identifier. The MHub.ai framework mediates the data flow using annotations and configurable workflows (**Figure 2b**). Every model is paired with *Structured Metadata*, which includes standardized documentation of the model architecture, intended use, training and test datasets, performance evaluation, and limitations, building on a standardized model card framework (Mitchell et al., 2019). We extend this definition with descriptors for input and output data, supported input modalities, if contrast-enhanced images were used during training, the recommended maximum slice thickness, license information, references to the original source code repository and publication, and example images.

The third component is the *Reproducibility Tests*, as each model contains public, real-world example inputs and reference output data stored in Zenodo (CERN, 2013) (**Figure 2c**). The MHub.ai framework test engine loads and processes the sample input data, compares the generated output with the reference data, and produces a machine- and human-readable report listing contextual differences, such as missing and extra files, Dice score for segmentation, and key-value comparison of structured files for prediction models. All MHub.ai models are tested as part of the contribution process (**see Appendix A, Section A7**) and before being published in the MHub.ai model repository.

**Model Repository:** We integrated 30 peer-reviewed, open-source AI models for segmentation (n = 20), prediction (n = 8), and feature extraction (n = 2) for CT (n = 18), MRI (n = 10), PET (n = 3), X-ray (n = 1), and pathology (n = 2) imaging modalities (**Figure 1b; Appendix C, Table C1**). Key models include *TotalSegmentator* for whole body anatomical segmentation (Wasserthal et al., 2023), *Platipy* for cardiac substructure segmentation (Finnegan et al., 2023), and *nnU-Net* based models for organ-specific tasks (Isensee et al., 2020). In oncology, the *Foundation Model for Cancer Imaging Biomarkers* enables self-supervised discovery of imaging biomarkers (Pai et al., 2024). For tumor segmentation in *PET/CT*, a leading model from the AutoPET challenge (Peng et al., 2023) is included. *LungMask* provides fast and accurate automated lung and lobe segmentation (Hofmanninger et al., 2020), while *PyRadiomics* supports quantitative feature extraction for predictive modeling (van Griethuysen et al., 2017).

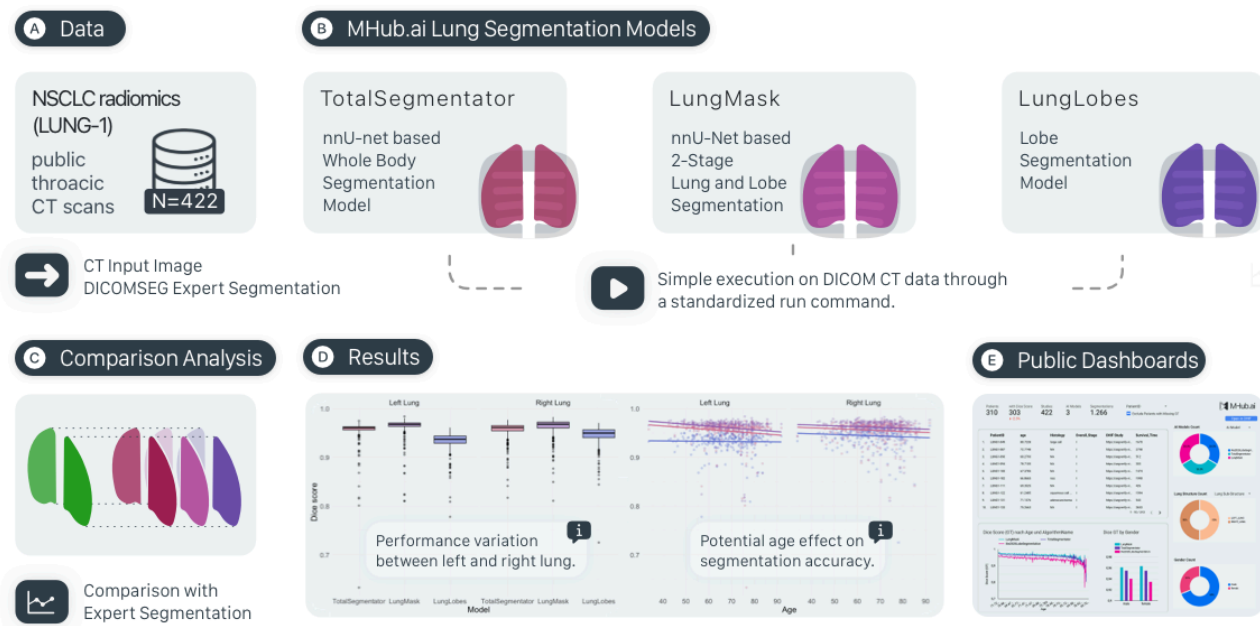

*Figure 3* Streamlined comparison of lung segmentation models.
a) *Public dataset input*: A total of 422 chest CT scans from the publicly available NSCLC-Radiomics dataset (Aerts et al., 2019) were processed, with expert segmentation available for 303 scans. b) *Unified Model Execution*: The three segmentation models were executed directly on the DICOM files using a standardized Docker run command, while the MHub.ai framework automatically performed DICOM preprocessing and model execution. c) *Comparative analysis*: The standardized segmentation results of the three models were compared with expert annotations using Dice similarity, for the right and left lung, to evaluate and compare the performance of the different models. d) *Insights into model differences*: Agreement between automated model segmentations and expert reference annotations is measured using the Dice score metric and shown by model for the left and right lungs (left), and plotted against age for each lung across models (right). In both views, Dice scores are high, with a stable relative ordering of models. e) *Public dashboards:* Results are made publicly available as interactive dashboards to recreate and extend our analysis results and perform visual case-level inspection entirely in the web browser.

**Use Case: Comparison of Lung Segmentation Models:** MHub.ai currently includes three independently developed lung segmentation models: TotalSegmentator (Wasserthal et al., 2023), LungLobes (Xie et al., 2020), and LungMask (Hofmanninger et al., 2020) (see **Figure 3b**). This use case demonstrates how MHub.ai enables standardized and reproducible comparison of independently developed AI models addressing the same clinical task (**Figure 3**).

We evaluated all three models on n = 422 chest CT scans from the publicly available NSCLC-Radiomics dataset (Aerts et al., 2019), with expert left–right lung annotations available for n = 303 patients. All models were executed directly on the DICOM files using the same MHub.ai execution command (**Appendix C, Figure C1**). The MHub.ai framework automatically handled DICOM preprocessing, model execution, and organization of the resulting segmentations. Lobe-level outputs were aggregated into left and right lung masks to match the expert reference annotations (**Appendix C**).

Segmentation performance was assessed using Dice similarity between each model and the expert annotation, calculated separately for the left and right lung. All models achieved high overall Dice scores. Small but statistically significant differences were found in certain stratifications: the LungLobes model showed slightly lower accuracy for the left lung ($\Delta$ = -0.0115, $p < .001$; see **Appendix C, Table C2).** A modest inverse correlation between Dice score accuracy and age was observed for LungMask (cor = -0.132, $p < .01$) and TotalSegmentator (cor = -0.116, $p < .01$; see **Appendix C, Table C3**). No clinically relevant differences were found between male and female patients (**Appendix C, Table C1**). Full statistical results are provided in **Appendix C**.

All segmentation outputs and derived metrics are publicly available through interactive dashboards (**Figure 3e; Appendix C, Figure C4**), enabling cohort-level summaries and case-level inspection of individual scans and segmentations.

By enabling direct execution on public clinical DICOM data, providing a harmonized application interface and generating DICOM compatible outputs, MHub.ai reduces the effort required to execute, compare, inspect, and reproduce analyses across multiple AI models.

**Integration with external Tools**: The harmonization of model implementations, inputs, outputs, and metadata also enables seamless integration into third-party systems. By implementing the MHub.ai

platform interface and using existing file standards, all models that comply with the proposed format become immediately available (see **Figure 1c**). We have developed MHubRunner (https://github.com/MHubAI/SlicerMHubRunner/), an extension for the open-source application 3D Slicer (Pieper et al., 2005). This extension uses our public web-based API to retrieve a list of MHub.ai models and provides users with a graphical user interface (GUI) to explore and run AI models, visually integrate the results, and directly apply them in clinical review workflows. Additionally, the standardized segmentation outputs and reference to the original image enable direct use within web-based viewers such as OHIF (Ziegler et al., 2020), supporting flexible visualization and integration in existing imaging environments.

**DISCUSSION**

MHub.ai offers a standardized framework for running AI model inference in medical imaging. Although many AI models are already publicly available in this field, few are widely adopted by third parties due to heterogeneous implementations, complex environment setups, limited support and documentation, and uncertain reproducibility. By combining model containerization, a standardized application interface, native support for clinical imaging data formats, structured model metadata, and automated reproducibility checks, MHub.ai defines a consistent distribution and execution format for trained AI models that lowers the barriers to model application and comparison on clinical imaging data.

For AI models to be reused by third parties, they must be easy to discover, set up, and execute. For model development and training, this has largely been addressed through widely adopted frameworks such as TensorFlow (Martín Abadi et al., 2015), PyTorch (Paszke et al., 2017), nnU-Net (Isensee et al., 2020), and MONAI (Cardoso et al., 2022), which provide standardized abstractions and interfaces. At inference time, however, a system-level approach is lacking: models differ in how they are packaged, in which format input images are provided, how inference is invoked, and how results are returned. While some models offer containerized distributions, these are not always available as ready-to-use images, and support for clinical data formats and outputs remains inconsistent. Existing model repositories, such as the MONAI Model Zoo, Hugging Face (Hugging Face, n.d.), BioImage Model Zoo (Ouyang et al., 2022), centralize model discovery but do not enforce standardized execution interfaces, uniform data handling, or structured and complete metadata and documentation.A notable example of an individual AI model that is used in external research projects is TotalSegmentator (Wasserthal et al., 2023), which has a straightforward installation process and a consistent inference interface for multiple segmentation tasks. However, these features are currently not systemic and remain specific to individual models.

To implement AI models in clinical or research workflows, their reliability must be evaluated. The accuracy of AI models is typically reported as metrics measured on internal or external datasets.

However, differences in the size and demographics of evaluation sets make it difficult to compare AI models solely based on reported metrics. Additionally, it is questionable how representative these metrics are for the specific use case in which the AI model will be implemented. It has been previously suggested to instead evaluate AI models recursively and internally (Youssef et al., 2023) on the target population. Consequently, evaluating AI models for a common task on the same dataset allows for direct performance comparison of model outputs under identical and, depending on data availability, controllable conditions. Lung segmentation on CT scans is one such task. Accurate segmentation of the lung on CT scans is critical for diagnosis, quantitative imaging, and radiation treatment planning, as precise delineation of the lung ensures effective dosing while sparing healthy tissue (Carmo et al., 2022, Hofmanninger et al., 2020). In our example use case evaluation, we demonstrate how the standardized interface of MHub.ai models enables simple, direct comparison of models and integration with public clinical data.

MHub.ai enforces a standardized structure for packaging and distributing medical imaging AI models at inference time. Model contributors must define an explicit end-to-end workflow, provide structured, machine-readable metadata, include representative example data, and pass automated reproducibility tests before models are included in the public model repository. This design ensures that models are delivered as self contained executable packages with a consistent application interface and native support for clinical imaging standards, defaulting to DICOM. As a result, models integrated into MHub.ai can be executed directly on clinical data without complex individual setup or time-consuming manual data conversion or organization. Additionally, the standardized output enables more consistent integration into research and clinical evaluation workflows.

The proposed approach and design decisions have also introduced some limitations to the platform in its current state. First, MHub.ai models are currently containerized using Docker, and our pre-built containers require the Docker engine to be installed on the host system. While Docker is widely used and offers a straightforward desktop application (Docker Desktop), it requires root access for installation and may be blocked by certain institutional policies. The MHub.ai framework operates independently of the containerization method, and the underlying principles of MHub.ai are not limited to Docker. They can be translated to alternative container runtimes such as Singularity (Kurtzer et al., 2017). Second, despite the outlined benefits, compliance with our format and guidelines, extensive metadata requirements, and reproducibility tests require additional effort from the model developer or a third-party contributor during model submission. To address this, we provide 18 reusable modules to reduce integration effort, along with step-by-step tutorials and detailed documentation. In addition, MHub.ai operates independently of and alongside other existing packaging approaches, reducing vendor lock-in and extending the reach of models through our repository. Third, current support does not cover all imaging modalities (e.g., whole slide pathology beyond TIFF), limiting applicability in some contexts. This can be added later to the platform because of its modular and extensible architecture. Fourth, the reproducibility tests are based on selected sample datasets that may not represent the full spectrum of clinical heterogeneity, which may limit the generalizability of the results. Fifth, most models require a GPU, either as a strict requirement or due to significant speed differences, which imposes a limitation on the host infrastructure.

While the proposed work provides a stable foundation, MHub.ai is an open platform specifically designed to be extensible for future improvements and developments. Extending format support to multiple platforms, including container runtimes such as Singularity, would improve compatibility with different computing environments. Integrating automatic quantification of performance metrics via embedded scoring modules would streamline end-to-end benchmarking. Implementing community-driven extensions through GitHub contributions, such as a domain-specific extension for pathology modules, would expand the coverage of reusable tools provided by the MHub.ai framework. Taken together, these developments would further lower barriers to adoption, improve reproducibility, and enable more reliable and scalable validation of AI models in medical imaging.

**METHODS**

We describe our AI model bundling strategy (see **Figure 4**), which includes the MHub.ai framework, metadata, reproducibility tests, the CI/CD pipeline to support platform contributions, and the dataset and model evaluation for the use case.

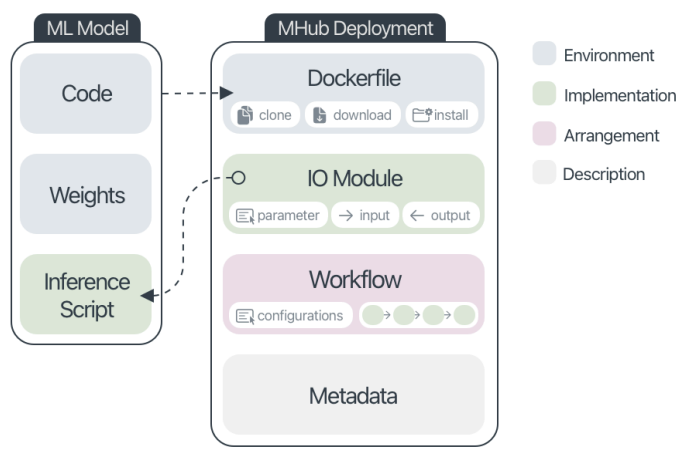

**Figure 4** The MHub.ai Model Deployment Format.
The MHub.ai deployment template is organized into environment, implementation, arrangement and description. System dependencies, model inference code and the trained weights are bundled in a container defined by a Dockerfile. The inference script is integrated through an IO Module adapter, including annotations for the required input data, a description of the output data and configurable parameters. The DICOM to DICOM execution is arranged in a workflow file, where configurable parameters and the sequential execution of IO Modules are defined. Model metadata is provided separately and in a defined schema.

**Containerization**

All models are containerized using Docker. Pre-built, ready-to-use container images are publicly available for all models on Docker Hub (repository: mhubai/model-name). We have defined a Dockerfile structure that includes the following steps: start from `mhubai/base:latest`, install the model dependencies, install the model source code, download the model weights, include the MHub.ai wrapper, and set the MHub.ai default entry point (see Environment in **Figure 4**). We provide a base image with common tools and our framework pre-installed and configured. The base image depends on the availability of CUDA GPU drivers on the host system that match the available hardware and are deliberately not included in the base image. Additional details are provided in **Appendix B1**.

**MHub.ai Framework**

The MHub.ai runtime environment is implemented as a Python package. Modularization and configurability are core concepts. Each task is implemented as an IO Module (see IO Module, **Figure 4**) and corresponds to an IO Module class. The framework provides Python decorators to conveniently annotate configuration parameters and input and output data, which are automatically resolved at runtime. The framework includes 18 base modules to support the processing of medical images: importers that search the input directory and import relevant files into a uniform semantic structure; filters that selectively refine datasets based on instance attributes or file availability; converters that convert images between formats such as NIFTI, MHA, TIFF, PNG, DICOMSEG, and RTStruct; exporters that compile and output metadata and custom reports; and organizers that orchestrate the selection and format of output. The AI inference code is called from a custom IO Module (see Implementation, **Figure 4**) that wraps the original AI model and defines the configurable parameters and input and output data. IO Modules are organized into workflows that are executed by the runtime engine. A workflow includes a label, description, sequential order of IO Modules, and configuration parameters in a single YAML file (see Workflow, **Figure 4**). For every MHub.ai model, the required default workflow operates on DICOM inputs and generates DICOM output where applicable (e.g., for all segmentation models). Additional details are provided in **Appendix B2-9**.

**Metadata Schema**

The model's metadata is provided in a single meta.json file (see Metadata, **Figure 4**). A JSON schema (Pezoa et al., 2016) defining the structure, required and optional keys, and value types of the file is publicly available in our repository and is automatically verified as part of our CI pipeline. Additional details are provided in **Appendix A4**.

**Reproducibility Testing**

Each model contains a sample input for each workflow, a representative input chosen specifically for the model. Wherever possible, we use real sample images from a public dataset. The model is run on the sample image to produce a reference output, which is stored on Zenodo (CERN, 2013) along with the sample input. As part of the MHub.ai framework, we provide a comparison tool with each model container that loads the sample and reference data, runs the model and outputs all results in a yaml report. The comparison checks are performed at directory level, where missing and additional files are listed, at file level, where differences in file size are recorded, and at content level depending on the file type. For segmentations, a Dice score below 0.99 compared to the reference segmentation is reported as a deviation. For structure files such as JSON or CSV, missing and additional key and value differences are reported. The report is created as a machine and human readable yaml file. As part of the contribution process, an independent reproducibility check is performed for each model. The results are publicly available on Github. The reproducibility tests are not limited to the provided sample output and reference output and can be used to ensure the reproducibility of studies and external projects. Additional details are provided in **Appendix B10**.

**CI/CD**

Continuous integration and continuous deployment (CI/CD) is implemented via Github Automations and ensures that every new model or update goes through automated format validation and metadata validation. We have set up an automated deployment pipeline that creates a model from a Pull-Request (PR), runs through the test pipeline and updates the pull requests with the test reports. Additional details are provided in **Appendix A7.**

**Datasets and Evaluation**

Our analysis includes a total of n = 422 patients diagnosed with non-small cell lung cancer (NSCLC) from the publicly available NSCLC radiomics dataset (Aerts et al., 2019). Each patient has a CT scan of the lung, with expert segmentation of the left and right lungs individually available for n = 303 patients. We downloaded the patient cohort as DICOM files from the Imaging Data Commons (IDC) archive using the Python utility idc-index (Fedorov et al., 2023). We ran three lung segmentation models available on MHub.ai (TotalSegmentator, LungMask, LungLobes) with the default run command directly on the downloaded data. We harmonized the output of models producing lobe-level segmentations by aggregating them into composite left and right lung masks to enable direct comparison with expert annotations using the standardized semantic labels provided by the MHub.ai framework. The LungMask model already provides separate whole-lung segmentation and requires no additional processing. Dice similarity is calculated pairwise between the segmentation generated by the model and the expert's segmentation. Additional details are provided in **Appendix C**.

**Software Availability**

- Code, workflows, and containers available at https://github.com/MHubAI.
- Web interface, model metadata, and documentation available at https://mhub.ai/.
- 3D Slicer plugin (MHubRunner) available via the Slicer extension manager and at https://github.com/MHubAI/SlicerMHubRunner.
- Interactive dashboards with lung segmentations, Dice scores, and clinical data available under https://lookerstudio.google.com/reporting/0c405365-31f5-4714-8b2a-03631e7a0686.

# Appendix A - Platform Design

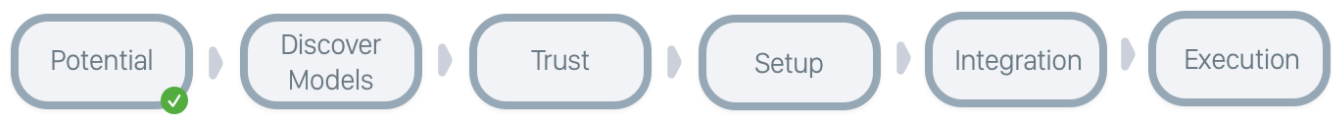

**Figure A1** Requirements to bridge the clinical implementation gap leading to the integration and usage of AI models.

## A1 - Bridging the Implementation Gap

The potential of AI models for clinical practice and medical research has been extensively documented in the literature (Hosny et al., 2018; Niazi et al., 2019; Huynh et al., 2020; Kann et al., 2021;McGenity et al., 2024; Mienye et al., 2025). However, only a limited number of models published in reputable journals and made available under permissive licenses have successfully made the transition to clinical application or have been adopted by research laboratories. Several factors contribute to this implementation gap. Previous studies have identified a lack of actionability, patient security and utility as main issues (Seneviratne et al., 2019) which can be summarized as a general lack of trust in AI models. The integration of AI can generally be understood as an investment into a new technology. While trust in AI models is often solely related to (patient) safety and linked to issues of privacy, model accuracy and stability, model generalizability and population bias, it takes confidence (trust in a positive outcome) to consider such an investment (see **Figure A1**). While several studies have identified the potential of AI models to reduce cost (Molwitz et al., 2025; van Leeuwen et al., 2022), the price of implementing AI models often remains unclear. Open source and open license models make the entire source code and the trained model weights available under a permissive license and can be used freely without license fees. This is required by some journals to increase transparency and emphasize open access and makes them *technically* available to external research departments and institutions. *Practically,* however, the implementation of such models still requires a lot of time, skill and experience and is generally not guaranteed to be successful.

It is therefore important to also look at the challenges and costs involved in setting up AI models in a hospital or research department. While there are additional investments to consider, e.g. investments in IT infrastructure, in storage and computing units such as GPU-equipped machines, or in the hiring of AI specialists, these are usually one-off expenses related to the investment in AI as a whole rather than an individual AI model. In this paper, we focus on the technical effort and challenges that make it difficult to set up and successfully embed AI models into clinical workflows or research pipelines, and propose a solution to reduce this effort, implicitly lowering the cost of integrating AI models and to help closing the implementation gap.

## A2 - Model Setup and Bundling App-like Models

AI models can be set-up remotely, e.g. using cloud infrastructure (Bontempi et al., 2023) or in a local environment. The setup entails the installation of the model and model dependencies, such as system libraries, required software tools and the pre-trained model weights. The complexity, potential obstacles and overall time investment required for this process varies greatly depending on the individual model requirements, the limitations and complexities of the local or remote IT infrastructure, the experience of the user and the quality of the documentation of the model.

AI models are typically written in scripting languages like Python rather than shipped as compiled executables and rely on specific system libraries and third-party tools (dependencies). In order to use them, all dependencies need to be made available (e.g., installed) on the system. The dependency version resolution, installation of dependencies, setup of file paths, provision of files, system configuration is referred to as *environment setup*. Recreating the exact environment setup is often challenging due to platform differences, dependency conflicts, and missing or outdated libraries, even for experienced developers.

A widely used strategy to solve this problem is *containerization*. A containerized runtime environment (*container*) provides a sandboxed environment that is separated from the host system environment and set up specifically for the ai model. The blueprint of a container is provided in the form of a static, immutable *image*. The environment setup is defined as a set of commands in a *Dockerfile*. Starting from a clean system environment (e.g., a specific version of Ubuntu), the commands of the Dockerfile are executed sequentially during the *build process* to derive the image. In this process, the AI model is configured and installed alongside all required dependencies. The derived image can easily be shared, e.g., as a compressed file or by uploading it to an image registry (e.g. Docker Hub). In contrast to the complex and system dependent setup, executing the image only requires the container engine to be installed and running on the host system.

Some AI models already provide Dockerfiles, outlying the steps of their environment setup and providing the blueprint to build an image. While this is a step forward, building an image from a Dockerfile itself is not guaranteed to succeed. Dependency conflicts and issues can still arise during the build process, e.g., due to patches and updates of any dependency (or a dependency of a dependency, …). Providing a self contained environment, the images are usually multiple GB large. Docker images mitigate this by creating a layer for every command and by sharing identical initial layers across images. Thereby, the total disk utilization can be reduced when two images start from the same base image and follow the same order of commands for their shared dependencies, however, such a synchronization of Dockerfiles requires a defined standard (e.g., provision of a unified base image) that all developers adhere to. Furthermore, there exists no standard for the environment setup within containers, including how or where input files are expected in the container and where the generated output files are located. It is also an often used practice for AI models to not include the large model weights as part of the code and to download them upon first execution. When containerizing AI models, this practice poses two problems. First, it enforces an internet connection at runtime. Second, since a container always starts from a fresh environment as defined in the image, the download occurs on every container startup, effectively adding to the execution time. Online dependencies should therefore be avoided and large files instead be downloaded during the build process and shipped as part of the image to keep them fully independent.

Comparing the installation requirements of a generic model and a containerized model (see **Figure A2**) shows that through containerization the individual preparation steps of the environment setup for the AI model are consolidated into the model image and the requirements are reduced to the container engine (e.g., Docker Desktop).

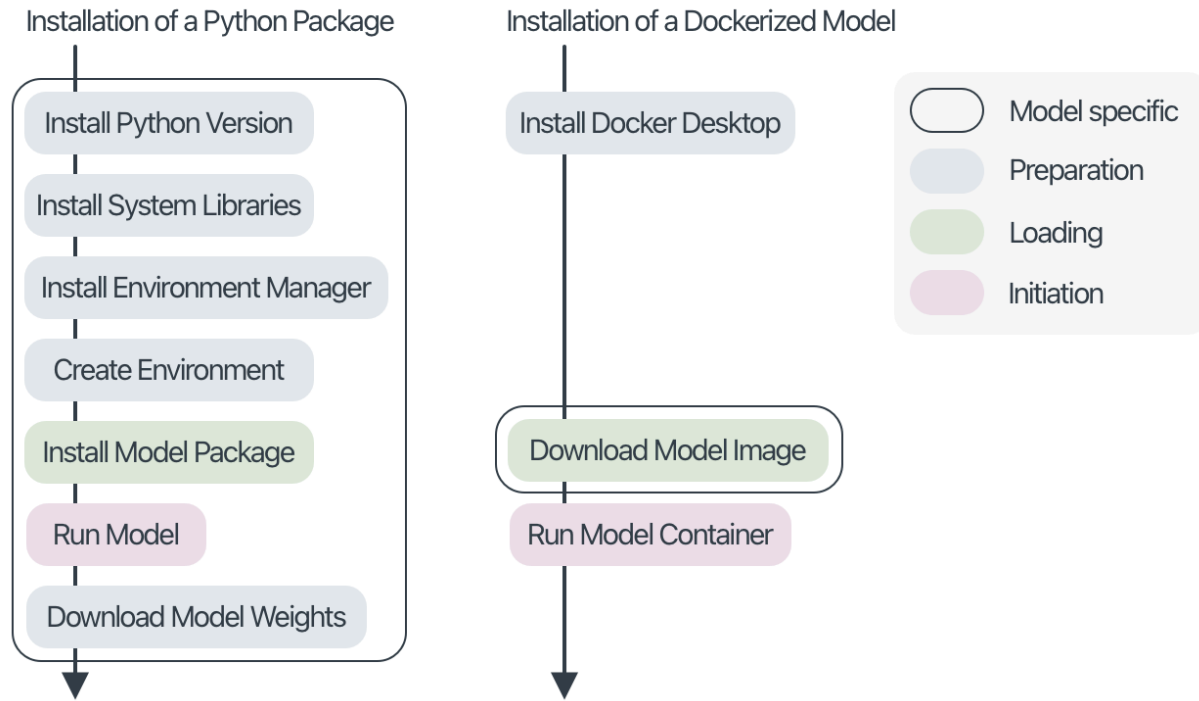

**Figure A2** Comparing the simplicity of preparation, loading, and initiation of a general AI model provided as a Python package with a containerized model. The outlined blocks are model-specific and must be adjusted depending on the model version. Containerization reduces the model specific installation steps.

MHub.ai uses containerization for AI models and provides a base image containing relevant tools and our framework that is shared between all models (see **Appendix B, Section B1**).

## A3 - Harmonization of Models

Bundling AI models into containers greatly simplifies setup and installation. The complex, model-specific environment setup becomes a one-time task of configuring the container engine. However, significant differences remain in the application interface of each model and in the model-specific requirements for data preparation before (input) and after (output) model execution. This results in a varied user experience across models and redundant effort in data organization, again a per-model investment (see **Figure A3**).

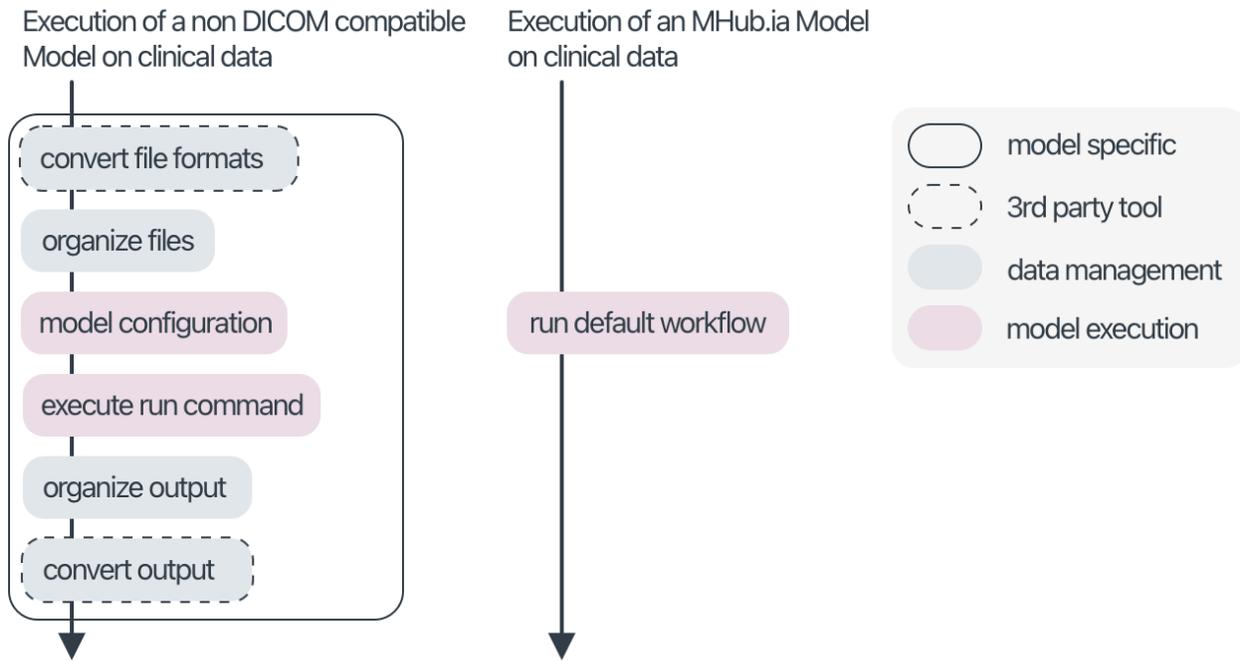

**Figure A3** Comparing the interaction through the application interface of a general AI model provided as a Python package with a containerized MHub.ai model, where the input data are clinical images in DICOM format and the AI model does not support DICOM inputs by default, requires tailored file conversion using third-party tools, manual organization of files into the required input format, and manual organization of the output to then convert it into a compatible DICOM format. In contrast, MHub.ai models run on DICOM and return DICOM by default, where applicable, without any individual model-specific adjustments required.

To solve this problem, we have developed the MHub.ai framework (**Appendix B, Section B2**) to harmonize the **input**, **output**, **configuration** and **execution** of the model. By providing a uniform standard application interface for all MHub.ai models, they can be easily integrated into existing workflows, support model substitution of similar models and enable local evaluation and model comparisons with reasonable effort.

In 1993, the Digital Imaging and Communications in Medicine (DICOM) standard was introduced for medical imaging (Clunie et al., 2000; Mildenberger et al., 2002). However, the data format is not ideal for training AI models and therefore many developers first convert the data to a more suitable data format. Therefore, most models expect a specific input format (usually the one used during training), while clinical reality is based on the DICOM standard. We offer conversion modules that are delivered as part of our base image and are bundled with the AI model. As a result, all MHub.ai models have a **uniform input interface** and run directly on DICOM data.

The format in which AI models output their calculations is usually also defined arbitrarily by the creators of the model, both semantically, e.g. how the output is to be interpreted (i.e. 100 percent = 1), and structurally, e.g. in which format the output is presented (i.e. 0 = background, 1 = heart). Both aspects need to be considered during the integration process of an AI model into an existing workflow. For segmentation models, RT-STRUCT and DICOM-SEG are two formats that extend the DICOM standard and refer to the original input image. The former is mainly used in radiotherapy for organs at risk and the delineation of tumors, but has limitations in the representation of complex shapes (Shrestha et al., 2024). DICOM-SEG is a newer format that supports the embedding of ROIs into semantic ontologies. For deep learning models that generate predictions, classifications or feature extractions, the most common file formats are JSON (more common in development environments) and CSV (usually used in clinics and for data analysis). In our MHub.ai framework, we have implemented abstract annotation of the output data and provide modules to convert segmentations to DICOMSEG or RTSTRUCT files and to generate JSON or CSV reports of clinical data. The selection and specification of the output format is defined in the workflow, resulting in a **uniform output interface**. The default workflow consistently generates a DICOMSEG for segmentation models.

By default, all MHub.ai models use a DICOM-based workflow. For advanced cases, different input formats, non-DICOM output, adaptation to local file structures, and other customizations can be provided as alternative workflows. The behavior of AI model pipelines is often controlled by additional parameters beyond the specification of input and output data. How these parameters are set is determined by the developers and varies between models. Within MHub.ai models, configurable parameters are exposed in a standardized way (see **Appendix B, Figure B3**) and are specified as part of the workflow configuration. While the default zero-setup workflow is simple and standardized, custom configurations can be provided as alternative workflows. Together, these features result in a **consistent execution interface and configuration**.

## A4 - Metadata and Repository

When selecting an AI model, it is important that the intended use, requirements and constraints are clearly defined and specified to avoid unexpected behavior. This includes information about acceptable input data, e.g. the modality (CT, MR, ...) that a model accepts, information about the semantics of the output to correctly interpret the results of the model, a description of additional parameters and configuration options. In the previous sections, we have explained how we provide a standardized interface for model setup, input, output, configuration and execution. In this section, we extend this with a standardized format for model metadata.

Mitchell et al. (2019) proposed the Model Card Framework to standardize the documentation of model architecture, intended use, training and test data, performance evaluation, and known limitations of algorithms. We extend this definition with optional elements such as images and descriptors for input and output data, which include the supported input modalities, compatibility with contrast-enhanced images, and the recommended maximum slice thickness. We also include relevant information such as the model's license, trained model weights, links to the model's original source code and publication, and citations of the original paper. The metadata for each model is provided in a single `meta.json` file. A JSON schema (Pezoa et al., 2016) is provided in our public repository to guarantee a consistent structure..

Based on the now comprehensive, standardized and machine-readable metadata, we can improve the discoverability (see **Figure A1**) and (pre-execution-) comparability of models in two ways. First, we have developed a public website with a model repository (https://mhub.ai/models) that contains all MHub.ai models and has a search and filter function that allows models to be filtered directly by key parameters such as the input modality or a specific region of interest in an output segmentation. In this way, the user can quickly search for "*all models run on a CT scan delineating the left upper lobe of the lung*". Second, we display the metadata for each model page in a consistent way so that the information is easily retrievable and one or more models can be compared side-by-side.

## A5 - Reproducibility

The complexity of AI model environments and pipelines not only makes it difficult to recreate the setup on a third party system. The reproducibility of AI models is also jeopardized by the reliance on rapidly evolving third-party libraries and tools (dependencies), such as data management and machine learning frameworks. Due to their active development and adaptation in a fast-moving field, they may introduce known or unknown changes alongside improvements and new features in upcoming releases. A setup installed today may not work in the future, even when following the same installation steps, unless the same version (and in the same order) is installed for each package - which sometimes not even the original developer can reproduce, as packages are often added and removed during the development process. One solution is to pin each dependency to an exact version. However, this only works if each dependency also binds all its dependencies to an exact version, which is not guaranteed and often outside of the control of the model developer. Moreover, this may not even be desirable. For example, many developers pin dependencies to smaller versions and allow patch updates for security-related updates. While this is a generally accepted and followed convention, an unexpected future update of a single dependency has the potential to break or change the output of an entire AI pipeline.

Containerization, as outlaid earlier, can solve this under specific circumstances. For instance, once built, a container image is a frozen snapshot of the entire AI model pipeline and all related dependencies. However, as soon as the image is built again, the above described situation applies, potentially allowing the model, its dependencies, the system environment or any other component not carefully fixed to change.

Since reproducibility is critical to building trust, model reproducibility was a major concern in the design and development of MHub.ai. Since we want to provide a consistent application interface for all models, constantly improve our MHub.ai framework and add new models to the platform, it is not practical to freeze models forever. Instead, we implement a mechanism in all MHub.ai models to ensure the model output stays consistent across image re-builds (see **Appendix B, Section B10**).

All MHub.ai models are tested as part of the contribution process (see **Section A7**) and sequentially before each build before they are published in the MHub.ai model repository (see **Appendix D**).

## A6 - Documentation and Tutorials

In the sections above, we have explained how the transfer of AI models to the MHub.ai format simplifies the setup and execution of these models and how the MHub.ai model repository facilitates the discovery and comparison of models based on structured metadata. To simplify the adoption of the proposed format, to help researchers and clinicians run MHub.ai models and integrate MHub.ai into new or existing workflows, and to help developers bring their own or third-party models into the proposed format, we provide extensive documentation, in which we explain the underlying concepts, how to execute and customize existing and create custom workflows, a detailed technical description of the MHub.ai framework, an overview of the existing modules and guidance for developers to create their own customized modules.

In addition, we have created four in-depth tutorials explaining step-by-step how to run a sample model on public DICOM data from the Imaging Data Commons (IDC) platform (*Run TotalSegmentator on IDC Collection*, **Table A1**), how to run a custom lung segmentation workflow on a non-DICOM format chest CT scan (*Run Custom MHub Lung*

*Segmentation Workflow on Chest CT in Nifti Format*, **Table A1**), how to deploy an open-source AI model in MHub.ai format (*Create the Thresholder Model for MHub*, **Table A1**), and how to switch the output of a cardiac segmentation model from DICOMSEG to RTStruct *(Run Heart Structure Segmentations as RTStruct*, **Table A1**). Most tutorials include a "too long didn't read" (TL;DR) section where the entire tutorial is summarized in a handful of commands to highlight the simplicity of our platform.

| Title | Duration | TL;DR |
| --- | --- | --- |
| [Run TotalSegmentator on IDC Collection](#) | 30 minutes | 6 commands |
| [Run Custom MHub Lung Segmentation Workflow on Chest CT in Nifti Format](#) | 45 minutes | 8 commands |
| [Create the Thresholder Model for MHub](#) | 90 Minutes | N/A |
| [Run Heart Structure Segmentations as RTStruct](#) | 15 minutes | 10 commands |

**Table A1** Overview of [MHub.ai](#) tutorials covering platform usage, adaptation and extension.

## A7 - Community Contributions

AI in medical imaging is a rapidly evolving field characterized by the continuous introduction of new tools, frameworks and trained models. The complexity of medical imaging with its various file formats, conversion tools and standards requires a high level of knowledge and experience. MHub.ai aims not only to simplify the application of AI, but also to provide developers with essential guidance, tools and support to provide their models in an accessible, versatile and easy-to-integrate format to amplify the impact of their work.

Our MHub.ai framework is written in Python, pre-installed in our base image and the driving engine behind all MHub.ai models. It provides semantic data annotation and queries, flexible workflow execution, automatic testing and modularization of workflow steps into building blocks (IO Modules). It has an extensive collection of 18 IO Modules for organizing and converting data into different formats. Custom IO Modules can be provided for each model, e.g. for model reference and optional processing steps.

We are aware of the rapid evolution of technology, where the frameworks used by AI engineers can become obsolete within a few years. For this reason, MHub.ai is designed to be framework-independent and does not interfere in the processes of model training and development. Instead, we focus on model inference and to provide a consistent interface and streamlined user experience. While MHub.ai provides the tools for the reproducible preparation of medical images, AI model development can be done independently of MHub.ai. Once a model has been trained and evaluated, it can be integrated into the MHub.ai format. It is important to note that this integration is not exclusive. MHub.ai does not have to be the only format or platform for model distribution and we support the parallel distribution in other platforms and formats.

MHub.ai is currently accepting model contributions via GitHub pull requests to our models repository. We have implemented GitHub automations that automatically validate the Dockerfile, JSON schemas and other relevant components. These automations also generate a comprehensive report that links to the relevant sections in the documentation.

## Appendix B - Platform Implementation

In the previous chapter, we described the prevailing challenges and justified our design decisions. In this chapter, we provide a comprehensive overview of the proposed MHub.ai model and the tools and templates we developed.

We define four layers (see **Figure 4**): environment, implementation, arrangement and metadata description. The environment layer includes the model setup and related dependencies. This layer covers the model code, trained weights, dependencies, system libraries, and configuration. The implementation layer provides the adapter between the AI model and the MHub.ai framework. Here, the configurable parameters, the required input data and the generated output data of the model are defined. The arrangement layer is where the AI model is embedded into a DICOM-to-DICOM workflow. This workflow consists of a sequential arrangement of IO Modules to import, convert, and prepare data, execute the model, and convert, link, and organize the output. Additionally, configurable parameter overrides are also defined within the workflow. Each MHub.ai model has a default workflow that is executed by default. The model metadata is defined in the description layer.

### B1 - Containerization Template

To support and guide the containerization of models while maintaining a shared structure among MHub.ai models, we provide a **base image** with pre-installed tools and minimal setup, and a **template** for building MHub.ai model images.

The base image includes the MHub.ai framework (see **Section B3**) and commonly used tools and libraries. Here we also define the universal file system endpoint for input and output data.

The template for MHub.ai model images specifies a set of mandatory rules for building MHub.ai model images that apply to the Dockerfiles:

1. All Dockerfiles must start from the latest MHub.ai base image:
   ```
   FROM mhubai/base:latest
   ```
2. Install and set up additional system libraries, tools and requirements.
3. The model definition must be imported using the provided buildutility:
   ```
   ARG MHUB_MODELS_REPO
   RUN buildutils/import_mhub_model.sh my_model ${MHUB_MODELS_REPO}
   ```
4. Additional collections can be imported using the provided buildutility:
   ```
   RUN buildutils/import_mhubio_collection.sh collection_name
   ```
5. Clone the model's public source code or install the model.
6. Download the model's weights.
7. The entrypoint must point to the MHub.ai run script (mhub.run)and the default workflow shall be executed by default.
   ```
   ENTRYPOINT ["mhub.run"]
   CMD ["--workflow", "default"]
   ```

Aligning all model images increases transparency within the container. It is also a fundamental design decision to provide a uniform application interface. The correct setup of the building instructions is automatically validated when a new pull request (PR) is created in the MHubAI/models repository as part of our continuous integration and continuous development pipelines (CI/CD) using GitHub automations.

While other containerization platforms are viable alternatives, we choose to initially focus our work on Docker due to its straightforward installation process and user-friendly graphical user interface (GUI). The described structure, however, can be translated to other containerization platforms. There are tools that allow the (semi-) automatic conversion to other platforms e.g. Apptainer (https://apptainer.org/).

## B2 - The MHub.ai Framework

Our framework is implemented as a Python package (https://github.com/MHubAI/mhubio) and pre-installed in our base image. It provides the core functionality for IO Modules (see **Section B3**) and workflow execution (see **Section B6**) and provides the uniform application interface for model execution (see **Section B7**) and model reproducibility tests (see **Section B10**).

## B3 - IO Modules

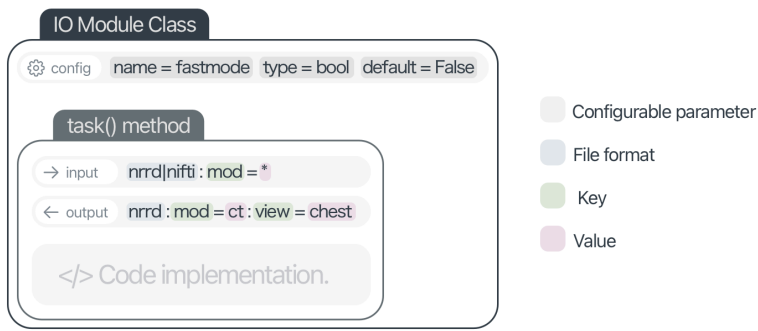

**Figure B1** Declarative Interface for Configurable Parameters and Semantic Input/Output Specifications.
The figure shows how IO modules declare configurable parameters, input descriptors, and output descriptors using framework-provided decorators. Semantic descriptors are used to query input files and define output files, and they are resolved into absolute file paths at runtime.

A core concept of MHub.ai is the separation of input-consuming (I), output-producing (O), and configurable (C) tasks or functionality into IO Modules that operate independently (see **Figure B1**). IO Modules are implemented as Python classes. The framework provides an abstract base class that is extended by specific IO Modules. Configurable parameters are implemented as class variables. The exposed name, data type, default value, and description are annotated with Python decorators provided by our framework. The task() method of an IO Module class contains the module implementation and is executed by the runtime engine. Required input files and generated output files are declared as parameters of the task() method, and Python decorators are used to semantically describe the input and output (see **Section B4**). Configuration, input, and output variables can be used within the task method and are automatically resolved by the runtime engine. The task method can contain complex Python implementations or, in its simplest form, run a CLI command, passing resolved file paths and configurations, acting as a simple adapter.

## B4 - Path Abstraction Through Semantic Annotation and Queries

Data exchange between IO Modules is managed by the data management layer of the framework. For each IO Module, the required input and generated output are declared through semantic queries and descriptors. A semantic descriptor registers a module's output with the runtime engine and includes a (file) type declaration followed by descriptive key-value pairs. A semantic query uses the same syntax but can include operators and placeholders. Within an IO Module, variables are provided for each input, output, and configurable parameter. The engine resolves these at runtime into absolute file paths and keeps track of related metadata in a local graph. It searches the graph for files matching the specified type and metadata, supplies the resolved files as module inputs, and generates absolute file paths for declared outputs. This mechanism creates a consistent directory structure within the container. Semantic declarations keep IO Modules universal and independent of custom naming conventions and directory layouts. The use of readable descriptors makes data flow easier to inspect and understand, and all intermediate files remain accessible and can be exported by specifying their semantic descriptors in the workflow configuration.

## B5 - IO Module Toolbox

Our framework offers 18 IO Modules, divided into seven functional categories to facilitate medical imaging workflows: Importers that standardize the ingestion of various data formats such as DICOM into a unified semantic structure; filters that selectively refine data sets based on instance attributes or file availability; converters that transform imaging data between formats such as NIFTI, MHA, TIFF, PNG, DICOMSEG and RTStruct without altering the underlying content; processors that extract or modify data while preserving its type, typically performing segmentation extractions and alignments; runners that execute AI pipelines to generate new analytics data; exporters that compile and output metadata and custom reports; and organizers that manage the arrangement and cleanup of files through flexible copy and delete operations. The external tools needed for these processes, e.g., dicomsort (Pieper, n.d.), plastimatch (Sharp et al., 2010), dcm2niix (Li et al., 2016), and itkimage2segimage (Herz et al., 2017), are pre-installed in our base image. Together, these modules support adaptable, transparent and reproducible data processing pipelines for medical imaging research and clinical applications.

## B6 - Workflows

Inside the container, the MHub.ai runtime engine sequentially executes IO Modules. Their order and configuration parameters define a workflow, represented by a single YAML file. Every model must include a default workflow, which is selected when not specified otherwise (see **Section B7**). The number of additional preconfigured workflows a model provides is not limited. Custom workflows can be passed as a string into the stdin of the run command. **Figure B2** illustrates an example default workflow. The first IO Module is a core module that scans the input directory for DICOM files and separates them into instances. The available instances are then, one by one, fed into the conversion module, here a DICOM to NIFTI converter. After this step, a NIFTI file is available for every input instance. The IO Module that follows is the custom inference adapter that runs the wrapped AI pipeline. The model requires NIFTI as the input format and requests this through semantic data annotation. As output, it defines a segmentation file, in NIFTI format, and the segmented ROI (see **Section B9**). The engine runs the ML Pipeline module for every available instance, resolving to the NIFTI file as input. When all instances are processed, another core IO Module converts the generated NIFTI segmentations into a DICOM-compatible output. As visualized by the arrows in the figure, it therefore requests two inputs: the original image in DICOM format and the segmentation in NIFTI format, and produces a DICOMSEG output file. After all instances are processed, the Export Module is run, organizing the files into the desired output structure. By default, for every instance, a subfolder with the SeriesInstanceUID is created in the output directory containing the segmentation file.

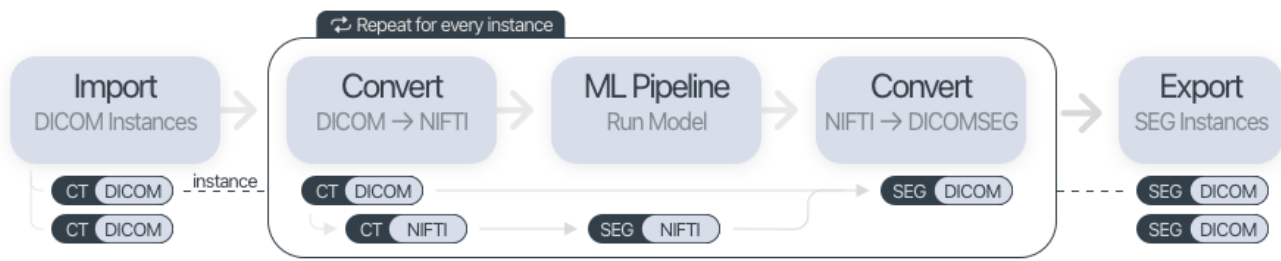

**Figure B2** Illustration of a typical MHub.ai DICOM-to-DICOM workflow.

The example process starts with a DICOM Import module that sorts separate files into individual instances. Next, a Conversion module converts the images into the required output format. The ML Pipeline, implemented as a custom module, runs the algorithm. A second Conversion module then converts the output generated by the pipeline, along with the original input image, into DICOMseg format. This process repeats for every instance identified by the Import module and is organized into an output file structure in the Export module**.**

## B7 - Execution and CLI Interface

All MHub.ai models share the same application interface. A docker run command creates the container from an MHub.ai model image and mounts the input and output directories. Because input and output mount points inside the container are standardized across MHub.ai models (see **Section B1**), the run command differs only in the model name (see **Figure B3a**). Inside the container, the run engine (mhub.run) provided by the MHub.ai framework is executed (**Figure B3b**). The run engine provides a CLI interface to specify the workflow (--workflow default), overwrite configurable parameters (--config:module.parameter=value), enable output logging and debugging. All cli parameters are optional and the default workflow is executed if no other workflow is specified. The run engine then executes the MHub.ai IO Modules in sequential order as defined in the workflow file (**Figure B2c**). The inference pipeline of the deployed AI model (**Figure B3d**) is called by the wrapper IO Module (m.Run) and provided with the required data.

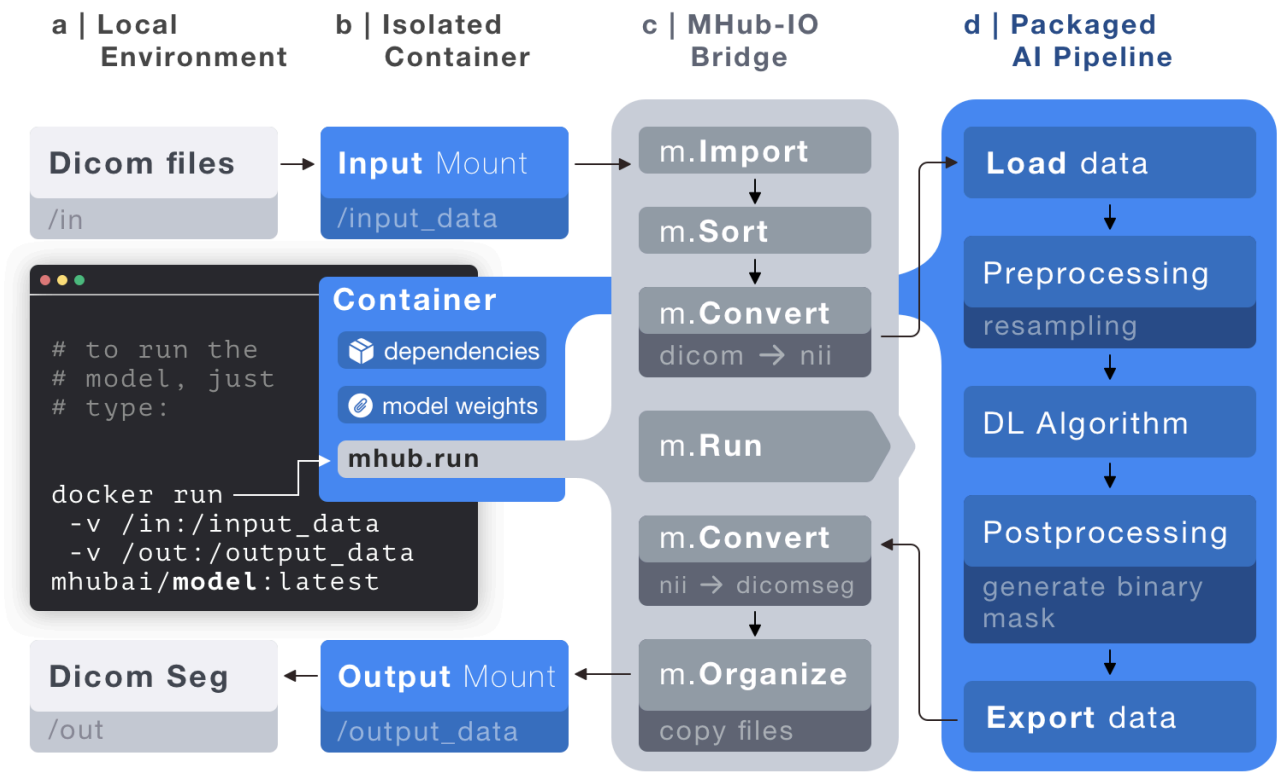

**Figure B3** Inside an MHub.ai container.

The black arrows indicate the data flow through the four layers. **a)** *Local environment*: The user specifies an input directory containing DICOM data and an output directory for storing results.. **b)** *Isolated container*: A Docker container encapsulates dependencies, model weights and the original AI pipeline and is executed with a standardized docker run command that mounts the input and output directories. **c)** *MHub.ai IO glue layer*: The MHub.ai execution framework orchestrates modules that import, sort, convert, run, and reorganize data. Semantic data types and declarative interfaces enable modules to function independently of the file system.. **d)** *Packaged AI pipeline:* The original AI pipeline is executed after data conversion and preparation by the MHub.ai framework.

## B8 - Extension API

Workflows are not limited to the included core IO Modules. Custom modules use the same interface and decorator conventions as standard modules. Custom modules can be directly included as part of the model definition. This is how the model's inference is called by implementing a wrapper IO Module that acts as adapter and enables the MHub.ai run engine to call the inference as part of a workflow. Such a custom module is only available to the specific model it is shipped with. For reusable IO Modules, the framework provides an extension API to include IO Modules from third party repositories at build time. When a workflow is loaded, the framework automatically discovers and registers all available modules (core modules, model-specific modules, and modules supplied through installed extension bundles) so they can be referenced by name in the workflow configuration file. Extension bundles must follow the required directory and module structure; bundles from unlisted repositories are supported but trigger a warning. An official list of supported extension bundles is available in the public model repository. This mechanism enables workflows to incorporate domain-specific or model-specific processing steps, supports reuse of common components, and allows functionality to be extended without modifying the underlying framework.

**B9 - SegDB**

A challenge in providing DICOM compatible output for segmentation models is the need for precise annotations. To help developers standardize segmentation labeling, we provide a keyword-based lookup table. We assign a unique identifier, e.g. HEART or LEFT_UPPER_LUNG_LOBE to each ROI and map this to a human-readable name, RGB color, and the semantic annotations that include category, type, and optional modifiers. For anatomical findings such as tumors, cysts or calcified plaques, semantic annotations are structurally different and require the specification of the context structure, e.g., the ROI where the anatomical finding relates to or is embedded in. We developed an intuitive annotation format, e.g. LEFT_LUNG+TUMOR, to indicate the presence of a neoplasm in the left lung and automatically resolve this into the semantic annotation structure required to generate DICOMSEG files at runtime. We bundled this functionality as a public python package ([https://github.com/MHubAI/SegDB/](https://github.com/MHubAI/SegDB/)) that is pre-installed in our base image.

**B10 - Test Framework**

The framework also includes a dedicated test engine that helps to verify whether a workflow produces reproducible outputs, e.g., when executed on third-party systems. Each model provides publicly accessible sample and reference data (e.g., via Zenodo) for this purpose. The test engine downloads these datasets, executes the workflow on the sample inputs, and compares the resulting outputs with the corresponding reference outputs. The comparison is summarized in a human- and machine-readable YAML report that records missing or additional files, differences in file size, and content-level discrepancies for supported file formats. For segmentation files, the engine computes Dice similarity scores per segment and flags scores below .99 compared to the reference segmentation in the report as deviation. For structured formats such as JSON or CSV, missing or extra keys and different values are reported on a key-value-pair level. The procedure is limited to assessing equivalence between generated and reference outputs and does not evaluate algorithmic performance beyond what can be inferred from these comparisons. In practice, the test engine can help detecting deviations caused by environment differences, dependency changes, or implementation errors, facilitates transparent and shareable reproducibility workflows in scientific studies, allows users to inspect expected outputs without executing the model, offers a standardized comparison method that avoids ambiguous manual inspection, and ensures consistent quality control before models are added to the public model repository.

## Appendix C - Use Case Lung Segmentation

We benchmarked three publicly available lung segmentation models integrated into MHub.ai, TotalSegmentator (Wasserthal et al., 2023), LungLobes (Xie et al., 2020), and LungMask (Hofmanninger et al., 2020), using a shared clinical cohort. First, we document standardized model execution (**Figure C1**), followed by cohort selection and preprocessing. Quantitative performance is then evaluated using Dice similarity metrics (**Figure C2b**), with further stratification by demographic variables (**Figure C3**). Finally, we describe how our results are made publicly available through interactive dashboards (**Figure C4**).

### C1 - Dataset and Model Characteristics

The evaluation cohort included 422 patients diagnosed with non–small cell lung cancer (NSCLC) from the publicly available NSCLC-Radiomics dataset (Aerts et al., 2019). The dataset includes a CT scan of the lung for each patient, as well as expert segmentations of the lung. For 309 patients, expert annotations are available as separate segmentations of the left and right lungs. For the remaining 113 patients, the expert annotation captures the whole-lung without left–right separation.

All three segmentation models generate lobe-level segmentations of the five lung lobes (upper and lower lobes in the left, upper, middle and lower lobe in the right lung). LungMask implements a two-stage segmentation pipeline in which the first stage generates segmentations of the entire left and right lung, which are included in the segmentation output alongside the lobe segmentation. The anatomical structures share the same annotation across all model outputs (see **Appendix B, Section B9**).

### C2 - Model Execution and Output Harmonization

All three lung segmentation models were executed using the same MHub.ai command-line interface. As shown in **Figure C1**, the execution command, input data declaration, and output directory specification are identical for all models; the only parameter that differs is the model identifier. The MHub.ai framework automatically identifies individual CT volumes, handles DICOM processing and conversion, executes the model on each image, and organizes the resulting segmentations into a standardized output structure. No manual data conversion or reorganization was required before model execution.

After execution, model outputs were harmonized to allow direct comparison with expert annotations. For TotalSegmentator and LungLobes, individual lobe segmentations were combined into composite left and right lung masks using the standardized semantic labels provided by the MHub.ai framework. LungMask outputs were already available at the whole-lung level and therefore did not require aggregation. This harmonization reconciled differences in output granularity without altering the anatomical evaluation target.

```
docker run -v /LUNG-1/dicom:/app/data/input_data:ro
           -v /LUNG-1/mhub:/app/data/output_data
              lungmask

docker run -v /LUNG-1/dicom:/app/data/input_data:ro
           -v /LUNG-1/mhub:/app/data/output_data
              gc_lunglobes

docker run -v /LUNG-1/dicom:/app/data/input_data:ro
           -v /LUNG-1/mhub:/app/data/output_data
              totalsegmentator
```

Input / Output
Model Name

**Figure C1** Execution commands for LungMask, LungLobe and TotalSegmentator model.
The run command, input data declaration and output folder specification is identical for all MHub.ai models. To select another model, only the model name needs to be changed in the run command.

### C3 - Cohort Exclusions and Eligibility for Quantitative Analysis

Quantitative evaluation using Dice similarity was performed only for cases where expert and model segmentations were spatially aligned and semantically compatible after output harmonization. Six patients were excluded due to processing errors caused by mismatches in image geometry between model outputs and expert annotations, preventing overlap computation. The 113 patients with expert annotation of the whole lung were excluded from the analysis. All remaining cases formed the basis for the statistical analyses reported below. The same filtered cohort of 303 patients was used consistently across all figures and tables.

### C4 - Quantitative Evaluation and Results

Segmentation accuracy was quantified using the Dice similarity coefficient, computed pairwise between each model's output and the corresponding expert lung segmentation. Dice scores were calculated separately for the left and right lungs.

**Figure C2** shows Dice score distributions stratified by model, lung side, and gender. All three models achieved high overall Dice scores, indicating strong agreement with expert annotations. The observed differences across demographics motivated additional statistical comparisons.

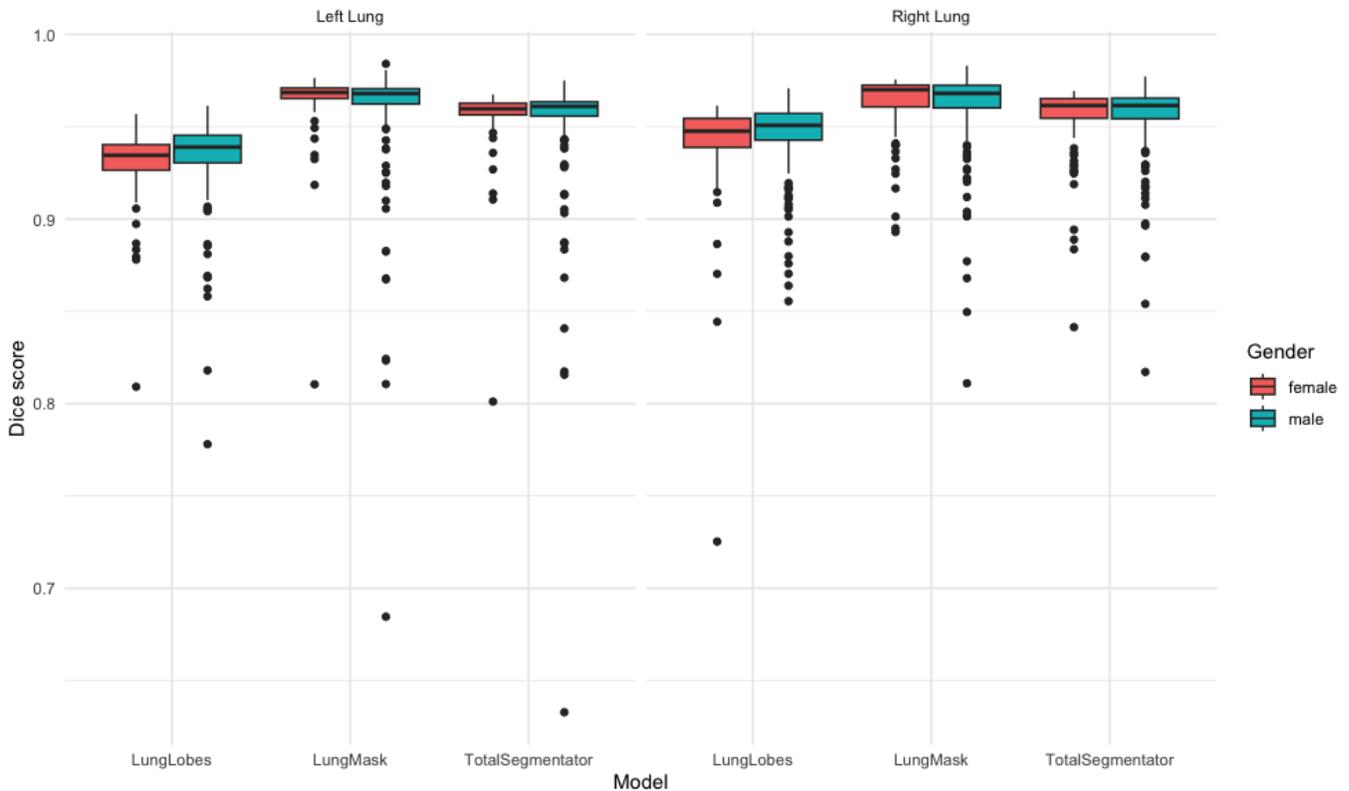

**Figure 2** Comparison of the Dice score (range .6 and 1, no outliers excluded) by model and gender for the prediction of left and right lung with the experts' segmentations.

First, we compared model accuracy between male and female patients. We performed an independent t-test comparing the Dice scores between model and expert annotations for male and female patients; the results are reported in **Table C1**. A statistically significant (p < .05) gender-associated difference (-0.00519) was observed for the LungLobes model, while no significant differences were found for LungMask or TotalSegmentator. Given the very small difference of less than .01 points in Dice score similarity, this difference is unlikely to be clinically relevant.

| Model | Estimate | p |
| --- | --- | --- |
| LungLobes | -0.00519 | p = .0171 |
| LungMask | 0.00267 | p = .155 |
| TotalSegmentator (v1) | 0.000478 | p = .807 |

**Table C1** t-tests comparing Dice scores between genders (male vs. female) per model. The differences in the mean values (estimate) and the corresponding p-values are given for each model.

In our second experiment, we compared model accuracy between the left and right lung. We performed independent t-tests across the three models; the results are shown in **Table C2**. We observed a small but statistically significant difference in accuracy between the left and right lung for the LungLobes model. Dice similarity is sensitive to target size, and because the left lung is anatomically smaller due to cardiac displacement, identical absolute segmentation errors can result in disproportionately lower Dice scores.

| Model | Estimate | p |
| --- | --- | --- |
| LungLobes | -0.0115 | p <.001 |
| LungMask | -0.000435 | p = .828 |
| TotalSegmentator (v1) | -0.00123 | p = .539 |

**Table C2** t-tests comparing Dice scores between lungs (left vs. right) per model. The differences in the mean values (estimate) and the corresponding p-values are given for each model.

Associations between Dice score and patient age are reported in **Table C3** and shown in **Figure C3**. LungMask and TotalSegmentator showed modest but statistically significant inverse correlations with age, while no significant association was found for LungLobes. Scatterplots show substantial overlap across age groups, indicating that age-related effects are small compared to overall inter-case variability.

| Model | $cor_s$ | $p_s$ |
| --- | --- | --- |
| LungLobes | 0.0128 | p = .756 |
| LungMask | -0.132 | p = .00134 |
| TotalSegmentator (v1) | -0.116 | p = .00485 |

**Table C3** Spearman rank correlations between Dice values and age by model. The table reports the Spearman rho ($cor_s$) and the corresponding p-values ($p_s$). The sample size (n = 590) is the same for all models.

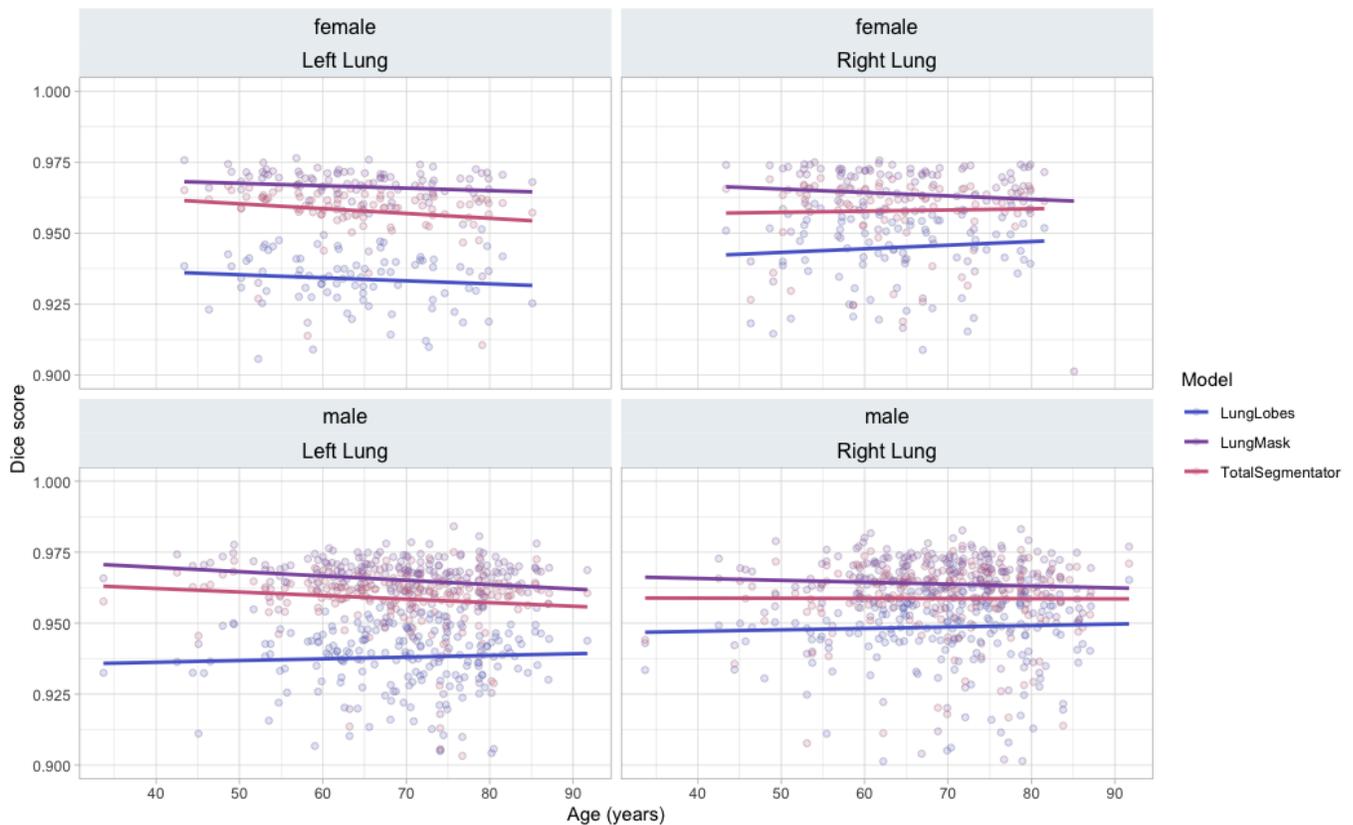

**Figure C3** Scatterplot of Dice scores (between .9 and 1) between segmentation and ground truth by age for each model, separated by lung (left, right) and gender (male, female).

## C5 - Reproducibility and Interactive Dashboards

To support transparency and reproducibility, the segmentation outputs and derived Dice scores are publicly available through interactive dashboards (**Figure C4**). These dashboards offer cohort-level summaries and case-level access to individual CT scans and corresponding segmentations connected to a cloud based OHIF (Ziegler et al., 2020) viewer. The dashboards are available at https://lookerstudio.google.com/reporting/0c405365-31f5-4714-8b2a-03631e7a0686.

All summary statistics shown in **Figures C2-C3** can be reproduced and traced back to individual cases using the dashboards, enabling inspection of outliers, failure modes, and demographic stratifications.

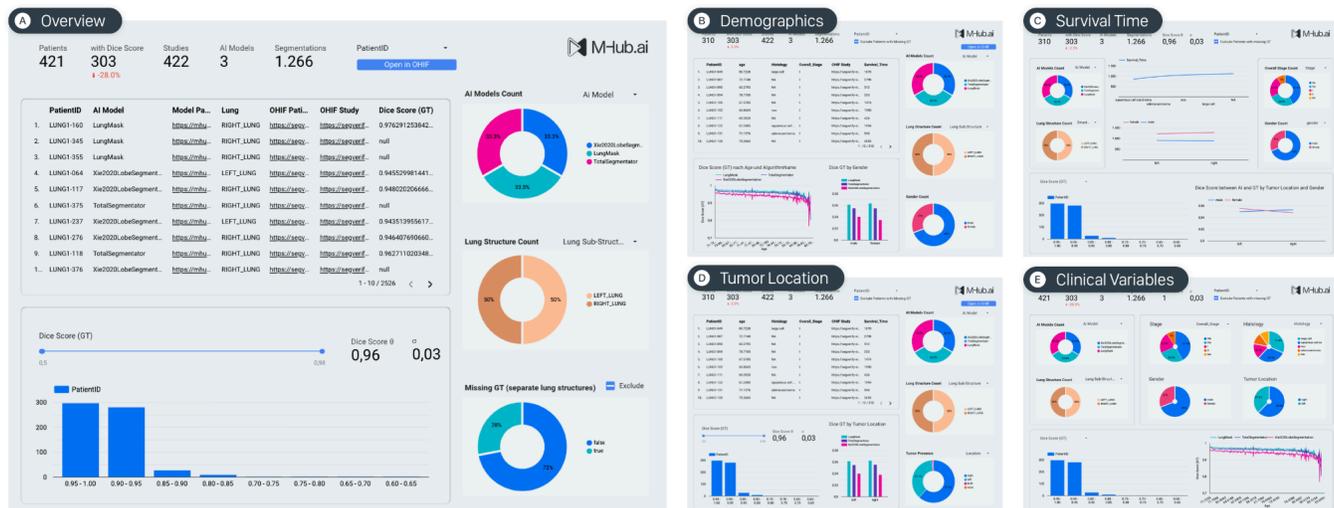

**Figure 4** Interactive dashboards supporting transparent and reproducible evaluation of lung segmentation models. The dashboards provide cohort summaries, performance metrics and case-level access to segmentation results and corresponding CT scans, enabling exploration across demographic and clinical variables. Panels A-E provide an overview of the model performance in general (A), separated by gender and age (B), insights into survival time (C), model performance variation under tumor presence (D), and separated by histology and overall cancer stage (E).

**Appendix D - Model Repository**

An overview of the resulting standardization across all 30 models can be found in **Table D1**. For each model, we collected information about the file format of the original input images of the model, the file format of the original output of the model, how model inference is executed through the application interface and the user manual (documentation) of the model.

As can be seen from the table (see column **A | Original**), 20/30 AI models expect a non-DICOM input format and require manual conversion of DICOM images to an alternative format used for inference. Five models implement custom logic to alternatively load a DICOM image from a folder, but require that the folder contains files for exactly one image, which may require a manual data organization step (e.g., sorting and separating DICOM files). Three models implement custom logic that allows image separation from mixed DICOM files, and one model has been published only in the MHub.ai format.

Only seven of the 20 segmentation models provide delineations with semantic tissue annotations in DICOM-compatible DICOMSEG format. All prediction models provide an individually and statically structured output format (mainly json, in one case also csv as an alternative format).

The application interface is a custom common line interface (CLI) for 11 models, five models are based on the nn-UNet model and use the same interface. 12 models require the input and output directories to be mounted at various locations into the Docker container, with nine models containing no additional configuration parameters. One model requires a custom metadata file to describe the input dataset and provides a separate script to create the file. For a model, the input and output data and other configurations are defined by the MONAI configuration file.

Less than 50% of all models (12/30) provide documentation with a complete description of the installation and setup process, the parameters of the application interface, address possible preparation and follow-up steps and give an example of how to use the model. Nine models provide a brief description or incomplete examples. No documentation is available for nine models.

| Model | T | A \| Original | | | | B \| MHub.ai | | | |
|---|---|---|---|---|---|---|---|---|---|
| | | Input Data | Output Format | Interface | Manual | Input Data | Output Format | Interface | Manual |
| **TotalSegmentator** (Wasserthal et al., 2023) | S | NIFTI | NIFTI | Custom CLI | Yes | DICOM | DICOMSEG | Workflow | Yes |
| **Platipy** (Finnegan et al., 2023) | S | NIFTI | NIFTI | Custom CLI | Yes | DICOM | DICOMSEG | Workflow | Yes |
| **BAMF MR Prostate Seg** (Murugesan et al., 2023) | S | NIFTI / DICOM | NIFTI / DICOMSEG | Docker Mount | Yes | DICOM | DICOMSEG | Workflow | Yes |
| **BAMF CT Kidney Seg** (Murugesan et al., 2023) | S | DICOM | DICOMSEG | Workflow | Yes | DICOM | DICOMSEG | Workflow | Yes |
| **CT Liver Seg** (Isensee et al., 2020) | S | NIFTI | NIFTI | nn-UNet CLI | Yes | DICOM | DICOMSEG | Workflow | Yes |
| **LungMask** (Hofmanninger et al., 2020) | S | NIFTI / DICOM | NIFTI | Custom CLI | Yes | DICOM | DICOMSEG | Workflow | Yes |
| **CaSuSt** (Nürnberg et al., 2022) | S | NIFTI / DICOM | NIFTI | Custom CLI | Yes | DICOM | DICOMSEG | Workflow | Yes |
| **Pulmonary Lobes Seg** (Xie et al., 2020) | S | MHA / DICOM | MHA | Custom CLI | No | DICOM | DICOMSEG | Workflow | Yes |
| **Pancreas Seg** Isensee et al., 2020 | S | NIFTI | NIFTI | nn-UNet CLI | Yes | DICOM | DICOMSEG | Workflow | Yes |
| **Tissue-Background segmentation in histopathological whole-slide images** (Bándi et al., 2019) | S | TIFF | MHA | Custom CLI + Config File | No | DICOM | TIFF | Workflow | Yes |
| **BAMF CT Liver Seg** (Murugesan et al., 2023) | S | NIFTI / DICOM | NIFTI / DICOMSEG | Docker Mount | Yes | DICOM | DICOMSEG | Workflow | Yes |
| **Second place in AutoPET challenge: False Positive Reduction Network** (Peng et al., 2023) | S | MHA | MHA | Docker Mount | No | DICOM | DICOMSEG | Workflow | Yes |
| **Whole Prostate Seg** (Isensee et al., 2020) | S | NIFTI | NIFTI | nn-UNet CLI | Yes | DICOM | DICOMSEG | Workflow | Yes |
| **Prostate Transitional and Peripheral Zone Seg** (Isensee et al., 2020) | S | NIFTI | NIFTI | nn-UNet CLI | Yes | DICOM | DICOMSEG | Workflow | Yes |
| **Foundation Model for Cancer Imaging Biomarkers** (Pai et al., 2024) | F | NIFTI | JSON | Custom CLI | Yes | DICOM | JSON | Workflow | Yes |
| **PI-CAI challenge baseline** (Bosma et al., 2023) | S | MHA | MHA | Docker Mount | No | DICOM | DICOMSEG | Workflow | Yes |
| **Thoracic OAR (nnU-Net)** (Isensee et al., 2020) | S | NIFTI | NIFTI | nn-UNet CLI | Yes | DICOM | DICOMSEG | Workflow | Yes |
| **Chest Radiograph Nodule Locator (Node21 Baseline)** (Sogancioglu et al., 2024) | P | MHA | JSON | Custom CLI + Docker Mount | Yes | DICOM | JSON | Workflow | Yes |
| **CT Lung cancer risk prediction** (Liao et al., 2019) | P | MHA | JSON | Docker Mount | No | DICOM | JSON | Workflow | Yes |

| Model | T | Input | Output | Config | Clinical Data | Input (MHub) | Output (MHub) | Config (MHub) | Clinical Data (MHub) |
|---|---|---|---|---|---|---|---|---|---|
| **Pancreatic Ductal Adenocarcinoma Detection in CT** (Alves et al., 2022) | S+P | MHA | MHA + JSON | Docker Mount | No | DICOM | DICOMSEG + JSON | Workflow | Yes |
| **PyRadiomics** (van Griethuysen et al., 2017) | F | NIFTI | JSON / CSV | Custom CLI + Config File | Yes | DICOM | JSON | Workflow | Yes |
| **Spine Seg (SPIDER Baseline)** (Lessmann et al., 2019) | S | MHA | MHA | Custom Metadata File | Yes | DICOM | DICOMSEG | Workflow | Yes |
| **BAMF MR Liver Seg** (Murugesan et al., 2023) | S | NIFTI / DICOM | NIFTI / DICOMSEG | Docker Mount | No | DICOM | DICOMSEG | Workflow | Yes |
| **Prostate Transitional and Peripheral Zone Seg (Prostate158)** (Adams et al., 2022) | S | NIFTI | NIFTI | Custom Monai Config | Yes | DICOM | DICOMSEG | Workflow | Yes |
| **TIGER challenge winner: Team VUNO** (VUNO, 2022) | P | TIF | JSON | Docker Mount | No | DICOM | JSON | Workflow | Yes |
| **STOIC2021 baseline** (Boulogne et al., 2023) | P | MHA | JSON | Docker Mount | Yes | DICOM | JSON | Workflow | Yes |
| **BAMF PET CT Breast** (Murugesan et al., 2023) | S | NIFTI / DICOM | DICOMSEG | Docker Mount | Yes | DICOM | DICOMSEG | Workflow | Yes |
| **SMIT Self-supervised Lung GTV Segmentation** (Jiang and Veeraraghavan, 2024) | S | NIFTI | NIFTI | Custom CLI | No | DICOM | DICOMSEG | Workflow | Yes |
| **FDG PET/CT Lung and Lung Tumor Annotation** (Murugesan et al., 2023) | S | NIFTI / DICOM | DICOMSEG | Docker Mount | Yes | DICOM | DICOMSEG | Workflow | Yes |
| **MRSegmentator** (Häntze et al., 2024) | S | NIFTI / MHA / NRRD / DICOM | DICOMSEG | Custom CLI | Yes | DICOM | DICOMSEG | Workflow | Yes |

**Table D1** Comparison of AI models in their original and MHub.ai deployment format.
Green = standardized / accessible, red = not standardized / not accessible, yellow = partially standardized / partially accessible. The model type (T) is S = segmentation, P = prediction, F = feature extraction.

After the implementation of all 30 models in the MHub.ai format, the effect of standardization becomes visible in the direct comparison of the before and after in the table (see column **B | MHub.ai**). All 30 models were enabled to run directly on clinical data, providing a standardized and clinically relevant input format. For 23/24 segmentation models, the output is provided as a standardized DICOMSEG file with consistent semantic tissue annotations, labels and color codes, and includes a link back to the original input Dicom image. One model uses pathology slide microscopy (SM), which is currently not directly supported by DICOMSEG. For all 4 prediction models, predictions are provided in the original, arbitrary format and additionally converted into semantic values that can be included in generic, fully workflow configurable reports. All 30 models are configured through workflow files that follow a defined and well-documented format. All 30 models share the same application interface, use the same execution command and rely on the same detailed documentation available for all MHub.ai models. In addition, all 30 models provide real sample data, a reference output and have passed our reproducibility test.

A key feature of the MHub.ai model repository is the overlap in segmentation outputs among models. Segmentation results are provided in a standardized output format, enabling direct pairwise comparison without additional data preparation or post-processing. Of the 30 integrated models, 17 (56.7%) share at least one segmentation output with another model (see **Table D2**). Combined with the low setup and execution cost, MHub.ai thereby supports systematic comparison and cross-evaluation of independently developed AI models.

| Number of Models with Output Overlap | Models | Share |
| --- | --- | --- |
| 0 | 13 | 43.3% |
| 1 | 5 | 16.7% |
| 2 | 1 | 3.3% |
| 3 | 3 | 10.0% |
| 4 | 6 | 20.0% |
| 10 | 1 | 3.3% |
| 12 | 1 | 3.3% |
| ≥1 | 17 | 56.7% |

**Table D2** Model-level overlap in the MHub.ai model repository.
Distribution of models that share at least one segmentation output with another model (n = 30 models).

# References

Adams, L. C., Makowski, M. R., Engel, G., Rattunde, M., Busch, F., Asbach, P., Niehues, S. M., Vinayahalingam, S., van Ginneken, B., Litjens, G., & Bressem, K. K. (2022). Prostate158 - An expert-annotated 3T MRI dataset and algorithm for prostate cancer detection. *Computers in Biology and Medicine*, *148*(105817), 105817.

Aerts, H. J. W. L., Wee, L., Rios Velazquez, E., Leijenaar, R. T. H., Parmar, C., Grossmann, P., Carvalho, S., Bussink, J., Monshouwer, R., Haibe-Kains, B., Rietveld, D., Hoebers, F., Rietbergen, M. M., Leemans, C. R., Dekker, A., Quackenbush, J., Gillies, R. J., & Lambin, P. (2019). *Data From NSCLC-Radiomics* [Data set]. The Cancer Imaging Archive. https://doi.org/10.7937/K9/TCIA.2015.PF0M9REI

Aggarwal, R., Sounderajah, V., Martin, G., Ting, D. S. W., Karthikesalingam, A., King, D., Ashrafian, H., & Darzi, A. (2021). Diagnostic accuracy of deep learning in medical imaging: a systematic review and meta-analysis. *Npj Digital Medicine*, *4*(1), 65.

Alves, N., Schuurmans, M., Litjens, G., Bosma, J. S., Hermans, J., & Huisman, H. (2022). Fully automatic deep learning framework for pancreatic ductal adenocarcinoma detection on computed tomography. *Cancers*, *14*(2), 376.

Bándi, P., Balkenhol, M., van Ginneken, B., van der Laak, J., & Litjens, G. (2019). Resolution-agnostic tissue segmentation in whole-slide histopathology images with convolutional neural networks. *PeerJ*, *7*(e8242), e8242.

Beam, A. L., Manrai, A. K., & Ghassemi, M. (2020). Challenges to the reproducibility of machine learning models in health care. *JAMA: The Journal of the American Medical Association*, *323*(4), 305–306.

Bontempi, D., Nuernberg, L., Pai, S., Krishnaswamy, D., Thiriveedhi, V., Hosny, A., Mak, R. H., Farahani, K., Kikinis, R., Fedorov, A., & Aerts, H. J. W. L. (2024). End-to-end reproducible AI pipelines in radiology using the cloud. *Nature Communications*, *15*(1), 6931.

Bosma, J. S., Saha, A., Hosseinzadeh, M., Slootweg, I., de Rooij, M., & Huisman, H. (2023). Semisupervised learning with report-guided pseudo labels for deep learning-based Prostate Cancer detection using biparametric MRI. *Radiology. Artificial Intelligence*, *5*(5), e230031.

Boulogne, L. H., Lorenz, J., Kienzle, D., Schon, R., Ludwig, K., Lienhart, R., Jegou, S., Li, G., Chen, C., Wang, Q., Shi, D., Maniparambil, M., Muller, D., Mertes, S., Schroter, N., Hellmann, F., Elia, M., Dirks, I., Bossa, M. N., … Dubois, M.-P. R. (2023). The STOIC2021 COVID-19 AI challenge: applying reusable training methodologies to private data. In *arXiv [eess.IV]*. https://doi.org/10.48550/ARXIV.2306.10484

Cardoso, M. J., Li, W., Brown, R., Ma, N., Kerfoot, E., Wang, Y., Murrey, B., Myronenko, A., Zhao, C., Yang, D., Nath, V., He, Y., Xu, Z., Hatamizadeh, A., Myronenko, A., Zhu, W., Liu, Y., Zheng, M., Tang, Y., … Feng, A. (2022). MONAI: An open-source framework for deep learning in healthcare. In *arXiv [cs.LG]*. arXiv. http://arxiv.org/abs/2211.02701

Carmo, D., Ribeiro, J., Dertkigil, S., Appenzeller, S., Lotufo, R., & Rittner, L. (2022). A systematic review of automated segmentation methods and public datasets for the lung and its lobes and findings on computed tomography images. *Yearbook of Medical Informatics*, *31*(1), 277–295.

Clunie, D. A. (2000). *DICOM structured reporting*. PixelMed Publishing.

European Organization for Nuclear Research, & OpenAIRE. (2013). *Zenodo*. CERN. https://doi.org/10.25495/7GXK-RD71

Fedorov, A., Longabaugh, W. J. R., Pot, D., Clunie, D. A., Pieper, S. D., Gibbs, D. L., Bridge, C., Herrmann, M. D., Homeyer, A., Lewis, R., Aerts, H. J. W. L., Krishnaswamy, D., Thiriveedhi, V. K., Ciausu, C., Schacherer, D. P., Bontempi, D., Pihl, T., Wagner, U., Farahani, K., … Kikinis, R. (2023). National Cancer Institute Imaging Data Commons: Toward transparency, reproducibility, and scalability in imaging artificial intelligence. *Radiographics: A Review Publication of the Radiological Society of North America, Inc*, *43*(12), e230180.

Finnegan, R. N., Chin, V., Chlap, P., Haidar, A., Otton, J., Dowling, J., Thwaites, D. I., Vinod, S. K., Delaney, G. P., & Holloway, L. (2023). Open-source, fully-automated hybrid cardiac substructure segmentation: development and optimisation. *Physical and Engineering Sciences in Medicine*, *46*(1), 377–393.

Häntze, H., Xu, L., Mertens, C. J., Dorfner, F. J., Donle, L., Busch, F., Kader, A., Ziegelmayer, S., Bayerl, N., Navab, N., Rueckert, D., Schnabel, J., Aerts, H. J., Truhn, D., Bamberg, F., Weiß, J., Schlett, C. L., Ringhof, S., Niendorf, T., … Bressem, K. K. (2024). MRSegmentator: Multi-Modality Segmentation of 40 Classes in MRI and CT. In *arXiv [eess.IV]*. arXiv. http://arxiv.org/abs/2405.06463

Herz, C., Fillion-Robin, J.-C., Onken, M., Riesmeier, J., Lasso, A., Pinter, C., Fichtinger, G., Pieper, S., Clunie, D., Kikinis, R., & Fedorov, A. (2017). Dcmqi: An open source library for standardized communication of quantitative image analysis results using DICOM. *Cancer Research*, *77*(21), e87–e90.

Hofmanninger, J., Prayer, F., Pan, J., Röhrich, S., Prosch, H., & Langs, G. (2020). Automatic lung segmentation in routine imaging is primarily a data diversity problem, not a methodology problem. *European Radiology Experimental*, *4*(1), 50.

Hosny, A., Parmar, C., Quackenbush, J., Schwartz, L. H., & Aerts, H. J. W. L. (2018). Artificial intelligence in radiology. *Nature Reviews. Cancer*, *18*(8), 500–510.

*Hugging Face*. (n.d.). Models. Retrieved January 8, 2026, from https://huggingface.co/

Huynh, E., Hosny, A., Guthier, C., Bitterman, D. S., Petit, S. F., Haas-Kogan, D. A., Kann, B., Aerts, H. J. W. L., & Mak, R. H. (2020). Artificial intelligence in radiation oncology. *Nature Reviews. Clinical Oncology*, *17*(12), 771–781.

Isensee, F., Jaeger, P. F., Kohl, S. A. A., Petersen, J., & Maier-Hein, K. H. (2021). nnU-Net: a self-configuring method for deep learning-based biomedical image segmentation. *Nature Methods*, *18*(2), 203–211.

Jiang, J., & Veeraraghavan, H. (2024). Self-supervised pretraining in the wild imparts image acquisition robustness to medical image transformers: an application to lung cancer segmentation. *Proceedings of Machine Learning Research*, *250*, 708–721.

Kann, B. H., Hosny, A., & Aerts, H. J. W. L. (2021). Artificial intelligence for clinical oncology. *Cancer Cell*, *39*(7), 916–927.

Koçak, B., Keleş, A., & Köse, F. (2024). Meta-research on reporting guidelines for artificial intelligence: are authors and reviewers encouraged enough in radiology, nuclear medicine, and medical imaging journals? *Diagnostic and Interventional Radiology (Ankara, Turkey)*, *30*(5), 291–298.

Korfiatis, P., Kline, T. L., Meyer, H. M., Khalid, S., Leiner, T., Loufek, B. T., Blezek, D., Vidal, D. E., Hartman, R. P., Joppa, L. J., Missert, A. D., Potretzke, T. A., Taubel, J. P., Tjelta, J. A., Callstrom, M. R., & Williamson, E. E. (2025). Implementing artificial intelligence algorithms in the radiology workflow: Challenges and considerations. *Mayo Clinic Proceedings. Digital Health*, *3*(1), 100188.

Kurtzer, G. M., Sochat, V., & Bauer, M. W. (2017). Singularity: Scientific containers for mobility of compute. *PloS One*, *12*(5), e0177459.

Lessmann, N., van Ginneken, B., de Jong, P. A., & Išgum, I. (2019). Iterative fully convolutional neural networks for automatic vertebra segmentation and identification. *Medical Image Analysis*, *53*, 142–155.

Li, X., Morgan, P. S., Ashburner, J., Smith, J., & Rorden, C. (2016). The first step for neuroimaging data analysis: DICOM to NIfTI conversion. *Journal of Neuroscience Methods*, *264*, 47–56.

Liao, F., Liang, M., Li, Z., Hu, X., & Song, S. (2019). Evaluate the malignancy of pulmonary nodules using the 3-D deep leaky noisy-OR network. *IEEE Transactions on Neural Networks and Learning Systems*, *30*(11), 3484–3495.

Liu, X., Faes, L., Kale, A. U., Wagner, S. K., Fu, D. J., Bruynseels, A., Mahendiran, T., Moraes, G., Shamdas, M., Kern, C., Ledsam, J. R., Schmid, M. K., Balaskas, K., Topol, E. J., Bachmann, L. M., Keane, P. A., & Denniston, A. K. (2019). A comparison of deep learning performance against health-care professionals in detecting diseases from medical imaging: a systematic review and meta-analysis. *The Lancet. Digital Health*, *1*(6), e271–e297.

Martín Abadi, Ashish Agarwal, Paul Barham, Eugene Brevdo, Zhifeng Chen, Craig Citro, Greg S. Corrado, Andy Davis, Jeffrey Dean, Matthieu Devin, Sanjay Ghemawat, Ian Goodfellow, Andrew Harp, Geoffrey Irving, Michael Isard, Jia, Y., Rafal Jozefowicz, Lukasz Kaiser, Manjunath Kudlur, … Xiaoqiang Zheng. (2015). *TensorFlow: Large-Scale Machine Learning on Heterogeneous Systems*. https://www.tensorflow.org/

McGenity, C., Clarke, E. L., Jennings, C., Matthews, G., Cartlidge, C., Freduah-Agyemang, H., Stocken, D. D., & Treanor, D. (2024). Artificial intelligence in digital pathology: a systematic review and meta-analysis of diagnostic test accuracy. *Npj Digital Medicine*, *7*(1), 114.

Mienye, I. D., Swart, T. G., Obaido, G., Jordan, M., & Ilono, P. (2025). Deep convolutional neural networks in medical image analysis: A review. *Information (Basel)*, *16*(3), 195.

Mildenberger, P., Eichelberg, M., & Martin, E. (2002). Introduction to the DICOM standard. *European Radiology*, *12*(4), 920–927.

Mitchell, M., Wu, S., Zaldivar, A., Barnes, P., Vasserman, L., Hutchinson, B., Spitzer, E., Raji, I. D., & Gebru, T. (2019). Model cards for model reporting. *Proceedings of the Conference on Fairness, Accountability, and Transparency*, 220–229.

Murugesan, G. K., McCrumb, D., Aboian, M., Verma, T., Soni, R., Memon, F., Farahani, K., Pei, L., Wagner, U., Fedorov, A. Y., Clunie, D., Moore, S., & Van Oss, J. (2023). AI-generated annotations dataset for diverse cancer radiology collections in NCI Image Data Commons. In *arXiv [eess.IV]*. arXiv. http://arxiv.org/abs/2310.14897

Najjar, R. (2023). Redefining radiology: A review of Artificial Intelligence integration in medical imaging. *Diagnostics (Basel, Switzerland)*, *13*(17), 2760.

Niazi, M. K. K., Parwani, A. V., & Gurcan, M. N. (2019). Digital pathology and artificial intelligence. *The Lancet Oncology*, *20*(5), e253–e261.

Nürnberg, L., Bontempi, D., \{de Ruysscher\}, D., Dekker, A., Hortal, E., Canters, R., & Traverso, A. (2022). Deep learning segmentation of heart substructures in radiotherapy treatment planning. *Physica Medica: European Journal of Medical Physics*, *104*(Supplement 1), S76.

Ouyang, W., Beuttenmueller, F., Gómez-de-Mariscal, E., Pape, C., Burke, T., Garcia-López-de-Haro, C., Russell, C., Moya-Sans, L., de-la-Torre-Gutiérrez, C., Schmidt, D., Kutra, D., Novikov, M., Weigert, M., Schmidt, U., Bankhead, P., Jacquemet, G., Sage, D., Henriques, R., Muñoz-Barrutia, A., … Kreshuk, A. (2022). BioImage Model Zoo: A community-driven resource for accessible deep learning in BioImage analysis. In *bioRxiv*. https://doi.org/10.1101/2022.06.07.495102

Pai, S., Bontempi, D., Hadzic, I., Prudente, V., Sokač, M., Chaunzwa, T. L., Bernatz, S., Hosny, A., Mak, R. H., Birkbak, N. J., & Aerts, H. J. W. L. (2024). Foundation model for cancer imaging biomarkers. *Nature Machine Intelligence*, *6*(3), 354–367.

Paszke, A., Gross, S., Chintala, S., Chanan, G., Yang, E., DeVito, Z., Lin, Z., Desmaison, A., Antiga, L., & Lerer, A. (2017). *Automatic differentiation in PyTorch*. https://openreview.net/pdf?id=BJJsrmfCZ

Peng, Y., Kim, J., Feng, D., & Bi, L. (2022). Automatic tumor segmentation via false positive reduction network for Whole-Body multi-modal PET/CT images. In *arXiv [eess.IV]*. arXiv. http://arxiv.org/abs/2209.07705

Pezoa, F., Reutter, J. L., Suarez, F., Ugarte, M., & Vrgoč, D. (2016, April 11). Foundations of JSON Schema. *Proceedings of the 25th International Conference on World Wide Web*. WWW '16: 25th International World Wide Web Conference, Montréal Québec Canada. https://doi.org/10.1145/2872427.2883029

Pieper, S., Halle, M., & Kikinis, R. (2005). 3D Slicer. *2004 2nd IEEE International Symposium on Biomedical Imaging: Macro to Nano (IEEE Cat No. 04EX821)*, 632-635 Vol. 1.

Pieper, Steve. (n.d.). *dicomsort: A project to provide custom sorting and renaming of dicom files*. Github. Retrieved January 8, 2026, from https://github.com/pieper/dicomsort

Seneviratne, M. G., Shah, N., & Chu, L. (2019). Bridging the implementation gap of machine learning in healthcare. *BMJ Innovations*, *6*, 45–47.

Sharp, G., Li, R., Wolfgang, J., Chen, G., Peroni, M., Spadea, M., Mori, S., Zhang, J., Shackleford, J., & Kandasamy, N. (05 2010). *PLASTIMATCH– AN OPEN SOURCE SOFTWARE SUITE FOR RADIOTHERAPY IMAGE PROCESSING*.

Shrestha, A., Watkins, A., Yousefirizi, F., Rahmim, A., & Uribe, C. F. (2024). RT-utils: A minimal Python library for RT-struct manipulation. In *arXiv [physics.med-ph]*. arXiv. http://arxiv.org/abs/2405.06184

Sogancioglu, E., van Ginneken, B., Behrendt, F., Bengs, M., Schlaefer, A., Radu, M., Xu, D., Sheng, K., Scalzo, F., Marcus, E., Papa, S., Teuwen, J., Scholten, E. T., Schalekamp, S., Hendrix, N., Jacobs, C., Hendrix, W., Sanchez, C. I., & Murphy, K. (2024). Nodule detection and generation on chest X-rays: NODE21 challenge. *IEEE Transactions on Medical Imaging*, *43*(8), 2839–2853.

Theriault-Lauzier, P., Cobin, D., Tastet, O., Langlais, E. L., Taji, B., Kang, G., Chong, A.-Y., So, D., Tang, A., Gichoya, J. W., Chandar, S., Déziel, P.-L., Hussin, J. G., Kadoury, S., & Avram, R. (2024). A responsible framework for applying artificial intelligence on medical images and signals at the Point of care: The PACS-AI platform. *The Canadian Journal of Cardiology*, *40*(10), 1828–1840.

van Griethuysen, J. J. M., Fedorov, A., Parmar, C., Hosny, A., Aucoin, N., Narayan, V., Beets-Tan, R. G. H., Fillion-Robin, J.-C., Pieper, S., & Aerts, H. J. W. L. (2017). Computational radiomics system to decode the radiographic phenotype. *Cancer Research*, *77*(21), e104–e107.

Vuno. (2022). *Vuno/tiger_challenge*. VUNO Inc. https://github.com/vuno/tiger_challenge/

Wasserthal, J., Breit, H.-C., Meyer, M. T., Pradella, M., Hinck, D., Sauter, A. W., Heye, T., Boll, D. T., Cyriac, J., Yang, S., Bach, M., & Segeroth, M. (2023). TotalSegmentator: Robust segmentation of 104 anatomic structures in CT images. *Radiology. Artificial Intelligence*, *5*(5), e230024.

Xie, W., Jacobs, C., Charbonnier, J.-P., & van Ginneken, B. (2020). Relational modeling for robust and efficient pulmonary lobe segmentation in CT scans. *IEEE Transactions on Medical Imaging*, *39*(8), 2664–2675.

Xu, Y., Li, Y., Wang, F., Zhang, Y., & Huang, D. (2025). Addressing the current challenges in the clinical application of AI-based Radiomics for cancer imaging. *Frontiers in Medicine*, *12*(1674397), 1674397.

Youssef, A., Pencina, M., Thakur, A., Zhu, T., Clifton, D., & Shah, N. H. (2023). External validation of AI models in health should be replaced with recurring local validation. *Nature Medicine*, *29*(11), 2686–2687.

Ziegler, E., Urban, T., Brown, D., Petts, J., Pieper, S. D., Lewis, R., Hafey, C., & Harris, G. J. (2020). Open Health Imaging Foundation Viewer: An extensible open-source framework for building Web-based imaging applications to support Cancer Research. *JCO Clinical Cancer Informatics*, *4*(4), 336–345.